\documentclass[10pt,journal,compsoc]{IEEEtran}

\ifCLASSOPTIONcompsoc
  \usepackage[nocompress]{cite}
\else
  \usepackage{cite}
\fi

\usepackage[backref,colorlinks=true]{hyperref}

\ifCLASSINFOpdf
  \usepackage[pdftex]{graphicx}
\else
  \usepackage[dvips]{graphicx}
\fi

\usepackage{amsmath}

\usepackage{algorithm}
\usepackage{algorithmic}

\usepackage{arydshln}

\usepackage[caption=false,font=footnotesize]{subfig}

\usepackage{url}

\usepackage{times}
\usepackage{ragged2e}
\usepackage{arydshln}
\usepackage{epsfig}
\usepackage{graphicx}
\usepackage[dvipsnames]{xcolor}
\usepackage{amsmath}
\usepackage{amssymb}
\usepackage{soul}
\usepackage{footnote}
\usepackage{scrextend}
\usepackage{amsmath}
\usepackage{xspace}
\usepackage{pifont}
\usepackage{booktabs}
\usepackage{multirow}
\usepackage{balance}
\usepackage{fancyvrb}
\usepackage[tableposition=top]{caption} %

\newcommand{\cmark}{\ding{51}}%
\newcommand{\xmark}{\ding{55}}%

\newcommand{\trainM}{\textsc{Sign-Train}$^{\text{M}}$\xspace}
\newcommand{\trainD}{\textsc{Sign-Train}$^{\text{D}}$\xspace}
\newcommand{\trainA}{\textsc{Sign-Train}$^{\text{A}}$\xspace}
\newcommand{\trainMD}{\textsc{Sign-Train}$^{\text{M,D}}$\xspace}
\newcommand{\trainMDA}{\textsc{Sign-Train}$^{\text{M,D,A}}$\xspace}
\newcommand{\valM}{\textsc{Sign-Val}$^{\text{M}}$\xspace}
\newcommand{\valD}{\textsc{Sign-Val}$^{\text{D}}$\xspace}
\newcommand{\valA}{\textsc{Sign-Val}$^{\text{A}}$\xspace}

\newcommand{\valMDA}{\textsc{Sign-Val}$^{\text{M,D,A}}$\xspace}
\newcommand{\testSign}{\textsc{Sign-Test}\xspace}

\newcommand{\SpottingFiltered}{Spotting-filtered\xspace}
\newcommand{\spottingFiltered}{spotting-filtered\xspace}
\newcommand{\trainSF}{\textsc{Sent-Train}$_{\textsc{SF}}$\xspace}
\newcommand{\valSF}{\textsc{Sent-Val}$_{\textsc{SF}}$\xspace}
\newcommand{\testSF}{\textsc{Sent-Test}$_{\textsc{SF}}$\xspace}

\newcommand{\trainManual}{\textsc{Sent-Train}$_{\textsc{H}}$\xspace}
\newcommand{\valManual}{\textsc{Sent-Val}$_{\textsc{H}}$\xspace}
\newcommand{\testManual}{\textsc{Sent-Test}\xspace}
\newcommand{\testManualNoSpace}{\textsc{Sent-Test}}

\newcommand{\datasetName}{BOBSL\xspace}
\newcommand{\bsldict}{BSLDict\xspace}
\newcommand{\bslonek}{BSL-1K\xspace}
\newcommand{\datasetNameLong}{BBC-Oxford British Sign Language\xspace}

\newcommand{\totalVocabSentences}{78K\xspace}

\newcommand{\totalSentencesApprox}{1.2M\xspace}
\newcommand{\totalSentencesPrecise}{1,193K\xspace}

\newcommand{\totalSentenceWordsPrecise}{11,356K\xspace}
\newcommand{\totalSigners}{39\xspace}
\newcommand{\totalHoursPrecise}{1,467\xspace}

\newcommand{\vocabTwoK}{2,281\xspace}
\newcommand{\vocabTwoKNoSpace}{2,281}

\newcommand{\numAnnosTwoKNoSpace}{452K}
\newcommand{\annosPerSignTwoK}{198\xspace}

\newcommand{\posetosign}{Pose$\rightarrow$Sign\xspace}
\newcommand{\videotosign}{I3D\xspace}

\newcommand{\wip}[1]{#1}

\newcommand{\wipTwo}[1]{#1}

\begin{document}
\title{BBC-Oxford British Sign Language Dataset}

\author{Samuel Albanie$^{*}$\quad\quad
        G\"ul Varol$^{*}$\thanks{* Equal contribution.}\quad\quad
        Liliane Momeni$^{*}$\quad\quad
        Hannah Bull$^{*}$\\
        Triantafyllos Afouras\quad\quad
        Himel Chowdhury\quad\quad
        Neil Fox\quad\quad
        Bencie Woll\\
        Rob Cooper\quad\quad
        Andrew McParland\quad\quad
        Andrew Zisserman%
\IEEEcompsocitemizethanks{
\IEEEcompsocthanksitem S.\ Albanie is affiliated with the Machine Intelligence Laboratory, University of Cambridge, UK, Email: sma71@cam.ac.uk
\IEEEcompsocthanksitem G.\ Varol is affiliated with LIGM, \'Ecole des Ponts, Univ Gustave Eiffel, CNRS, France
\IEEEcompsocthanksitem L.\ Momeni, T.\ Afouras, H.\ Chowdhury and A.\ Zisserman are affiliated with the Visual Geometry Group, University of Oxford, UK
\IEEEcompsocthanksitem H. Bull is affiliated with LISN, Univ Paris-Saclay, CNRS, France
\IEEEcompsocthanksitem N.\ Fox and B.\ Woll are affiliated with the Deafness, Cognition and Language Research Centre, University College London, UK
\IEEEcompsocthanksitem R.\ Cooper and A.\ McParland are affiliated with BBC Research \& Development, London, UK

}
}

\markboth{
    Albanie \MakeLowercase{\textit{et al.}}: BBC-Oxford British Sign Language Dataset}{
    Albanie \MakeLowercase{\textit{et al.}}: BBC-Oxford British Sign Language Dataset
}

\IEEEtitleabstractindextext{%
\begin{abstract}
In this work, we introduce the
BBC-Oxford British Sign Language (BOBSL) dataset,
a large-scale video collection of British Sign Language (BSL).
BOBSL is an extended and publicly released dataset based on the BSL-1K dataset~\cite{Albanie2020bsl1k}
introduced in previous work.
We describe the motivation for the dataset,
together with statistics and available annotations.
We conduct experiments to provide baselines
for the tasks of sign recognition, sign language alignment,
and sign language translation.
Finally, we describe several strengths and limitations
of the data from the perspectives of machine learning
and linguistics,
note sources of bias present in the dataset,
and discuss potential applications of \datasetName
in the context of
sign language technology.
The dataset is available at~\url{https://www.robots.ox.ac.uk/~vgg/data/bobsl/}.
\end{abstract}

}

\maketitle
\IEEEpeerreviewmaketitle
\IEEEdisplaynontitleabstractindextext
{
  \hypersetup{hidelinks}
  \tableofcontents
}
\clearpage

\section{Introduction}\label{sec:intro}

Sign languages are visual languages that have evolved
in deaf communities.
They possess complex grammatical structures
and lexicons~\cite{sutton1999linguistics},
akin to the complexity of spoken languages.
In this paper, we introduce a large-scale dataset of
British Sign Language (BSL), the sign language of the
British deaf community.

To date, a central challenge in conducting sign language
technology research  has been a lack of large-scale public
datasets for training and evaluating computational 
models~\cite{bragg2019sign}.
The goal of the \datasetNameLong (\datasetName) dataset
is to provide a
collection of BSL videos to support 
research on tasks such as
sign recognition, sign language alignment and sign language translation.

The rest of the paper is structured as follows:
in Sec.~\ref{sec:dataset} we provide an overview of
the \datasetName dataset;
in Sec.~\ref{sec:dataset-collection}, we describe the
collection and annotation (both automatic and manual) of the dataset, and also the evaluation partitions.
Next, in Sec.~\ref{sec:models} we give implementation
details and descriptions of models for baselines on the tasks of recognition, alignment and translation. In
Sec.~\ref{sec:experiments} we present our evaluation
protocols and baseline results for the dataset.
In Sec.~\ref{sec:implications} we discuss
the opportunities and limitations of the data
from the perspectives of sign linguistics
and downstream applications
and note several sources of bias present in the data
before 
concluding in Sec.~\ref{sec:conclusion}.
\section{\datasetName Dataset Overview}\label{sec:dataset}

In this section, we first give an overview of
\datasetName content and statistics (Sec.~\ref{subsec:content}).
Next, we compare BOBSL to existing sign language datasets
(Sec.~\ref{subsec:prior}), 
outline data usage terms (Sec.~\ref{subsec:usage}) and
describe its relationship to the
\bslonek dataset (Sec.~\ref{subsec:relationship-to-bslonek}).

\subsection{Dataset content and statistics}
\label{subsec:content}

\begin{figure}[t!]
    \centering
    \includegraphics[width=0.48\textwidth]{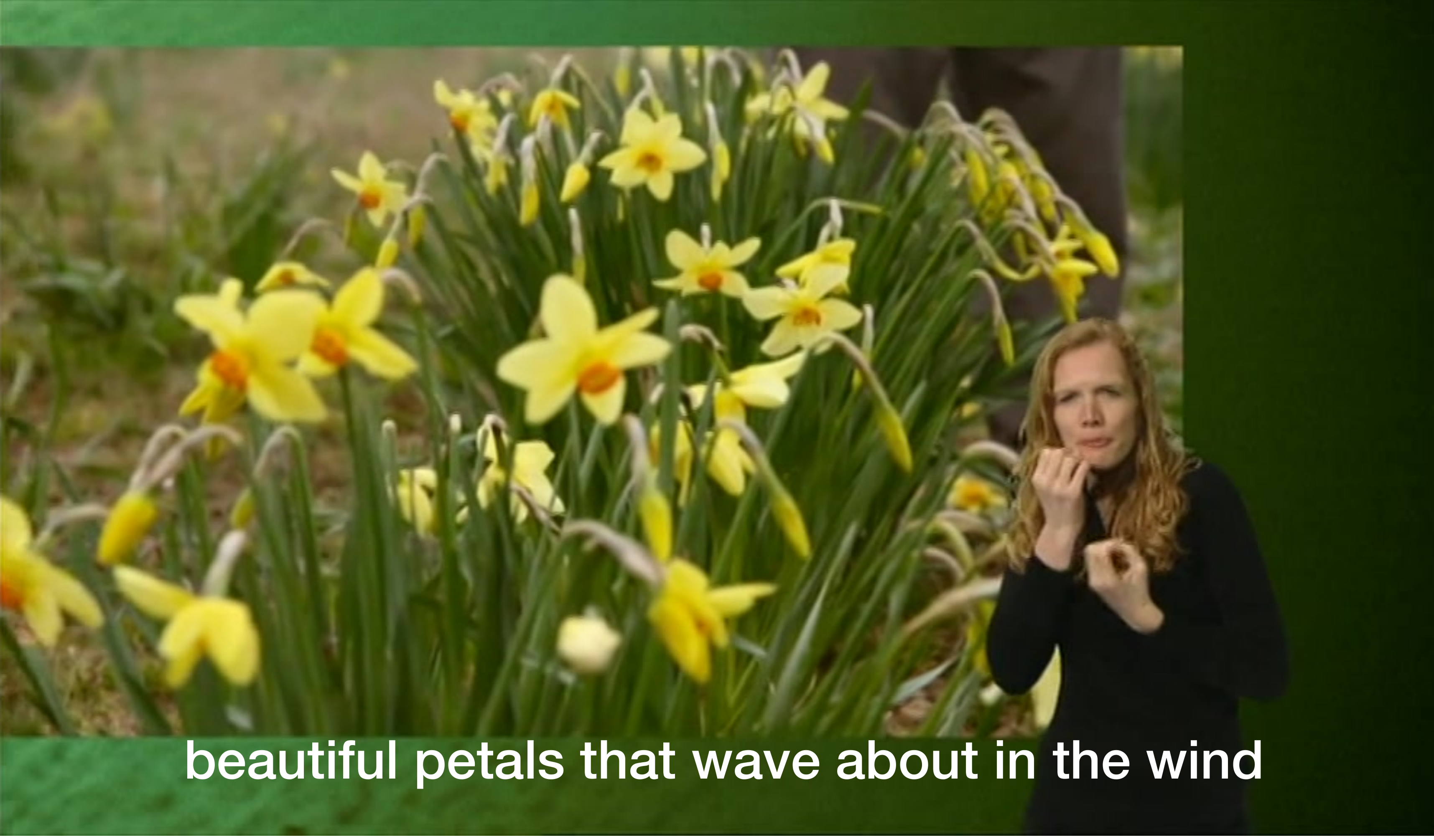}
    \caption{
    \textbf{BOBSL source data.} The source data consists of British Sign Language interpreted footage of BBC broadcasts (in this example from a \textit{ Gardeners' World} program), along with English subtitles corresponding to the audio content. 
    }
   
    \label{fig:example_with_sub}
\end{figure}

\begin{figure}[t!]
    \centering
    \includegraphics[width=0.48\textwidth]{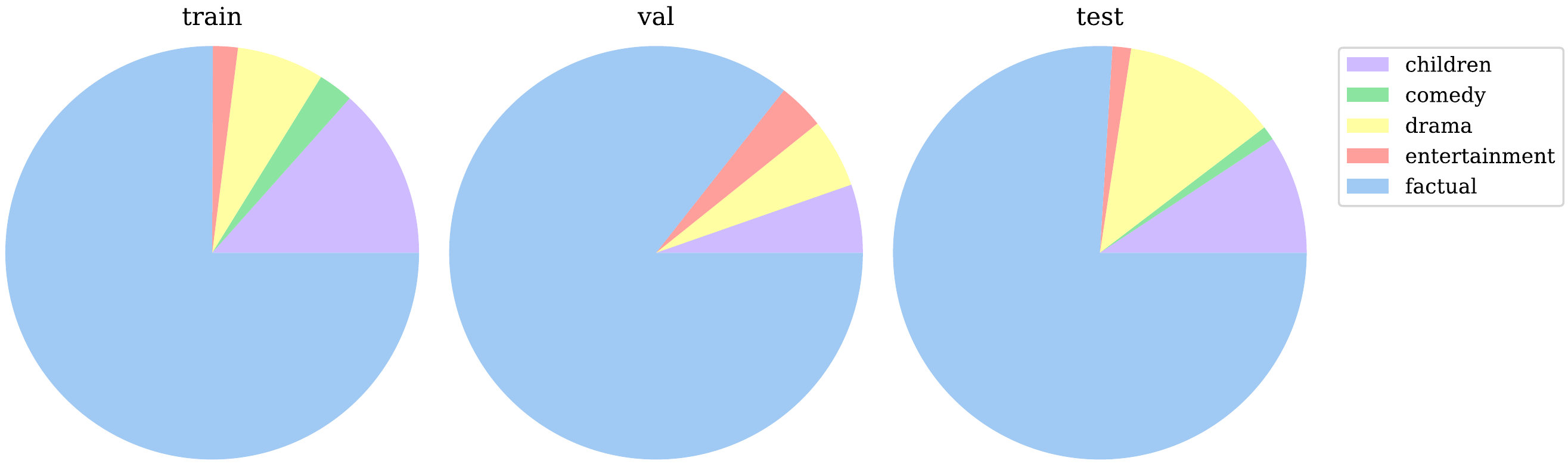}
    \caption{
    \textbf{BOBSL division into genres.} The duration of each BOBSL dataset split can be divided into 5 genres, with \textit{factual} representing the largest proportion for train, validation and test splits. 
    }
    \label{fig:pie_charts_genre}
\end{figure}

\begin{figure}[t!]
    \centering
    \includegraphics[width=0.48\textwidth]{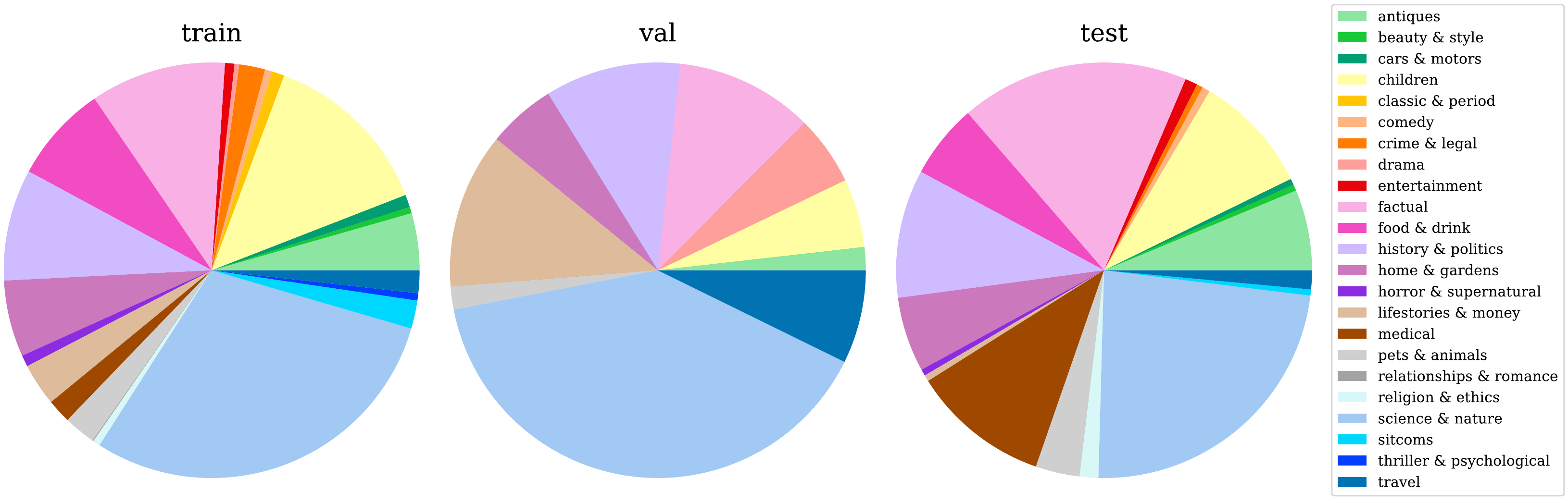}
    \caption{
   \textbf{BOBSL division into topics.} Each BOBSL dataset split can be divided into 22 topics, with \textit{science \& nature} representing the largest proportion for train, validation and test splits.
   The figure is best seen on computer screen and in colour.
    }
    \label{fig:pie_charts_topic}
\end{figure}

The data consists of BSL-interpreted BBC broadcast footage,
along with English subtitles corresponding to the audio content,
as shown in Fig~\ref{fig:example_with_sub}.
The data contains 1,962 \textit{episodes},
which span a total of 426 differently named TV \textit{shows}.
We use the term \textit{episode} to refer to a single video of contiguous broadcast content, whereas a show (such as \textit{``Countryfile''}) refers to a collection of episodes grouped thematically by the broadcaster,
whose episodes typically share significant overlap in subject matter, presenters, actors or storylines.
The shows can be partitioned into five genres using
BBC metadata as shown in Fig.~\ref{fig:pie_charts_genre};
with the majority of shows being \textit{factual}, i.e.\ documentaries.
These can be further divided into 22 topics,
as shown in Fig.~\ref{fig:pie_charts_topic}. 
Including horror, period and medical dramas, history, nature and science documentaries,
sitcoms, children's shows, and programs covering cooking, beauty, business and travel,
the BOBSL data covers a wide range of topics.

\begin{figure}
    \centering
    \includegraphics[trim={3cm 0.5cm 4cm 1cm},clip,width=0.48\textwidth]{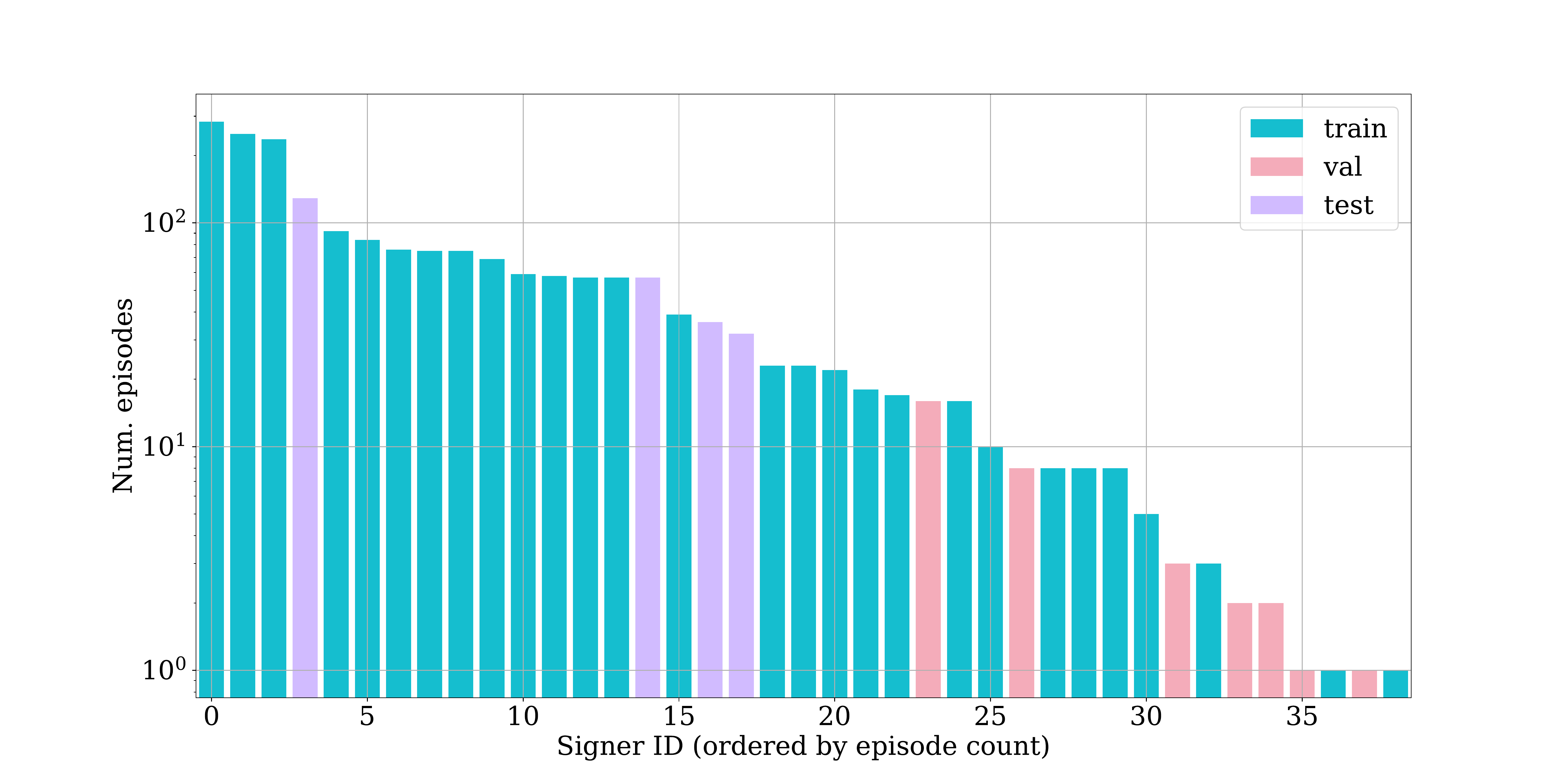}
    \caption{
    \textbf{Distribution over signers.}
    The number of episodes associated with each BSL interpreter in the
    \datasetName dataset follows a power law distribution (note the log-scale on the y-axis). 
    }
    \label{fig:signer-dist}
\end{figure}

\begin{figure}
    \centering
    \includegraphics[trim={0cm 0cm 0cm 0cm},clip,width=0.48\textwidth]{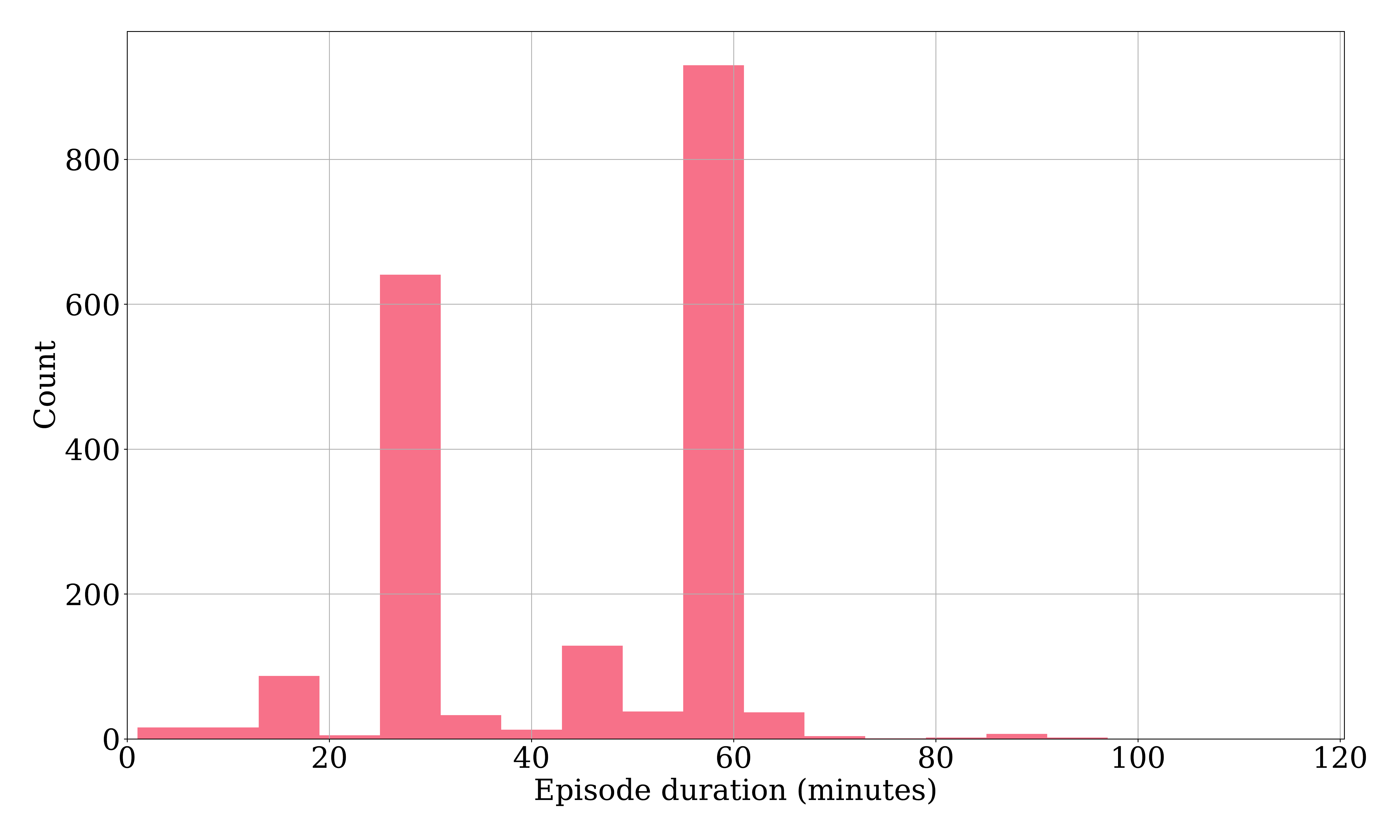}
    \caption{
    \textbf{Distribution over episode durations.}
    The duration of episode videos in the \datasetName dataset.
    The majority of episodes are either 30 minutes or 60 minutes in duration,
    with the longest episode lasting 120 minutes.
    }
    \label{fig:episode-durations}
\end{figure}

Statistics of the BOBSL data are presented in Tab.~\ref{tab:splits}.
The 1,962 episodes have a duration of approximately \totalHoursPrecise hours
(i.e.\ 45 minutes per episode on average,
with the majority of episodes lasting approximately 30 or 60 minutes,
as shown in Fig.~\ref{fig:episode-durations}).
The videos have a resolution of $444 \times 444$ pixels and a frame rate of 25 fps.
There are approximately \totalSentencesApprox sentences extracted from
English subtitles covering a total vocabulary size of \totalVocabSentences English words.
BOBSL contains a total of \totalSigners signers (interpreters).
We divide the data into train, validation and test
splits based on signers, to enable signer-independent evaluation, i.e.\ there is no signer overlap between
the three splits.
The distribution of programs associated to each signer,
together with the split information is illustrated in Fig.~\ref{fig:signer-dist}. We note that a few signers appear very frequently.


\begin{table*}
    \setlength{\tabcolsep}{6pt}
    \centering
    \caption{
    \textbf{Statistics summarising the data distributed
    across the splits of \datasetName.}
    \textit{Num. Signers} indicates the number of signer
    identities within a partition,
    \textit{Num. Raw Subtitles} denotes the number of subtitles
    (which do not necessarily form complete sentences)
    associated with the original broadcasts,
    while  \textit{Num. Sentences} indicates the number of English sentences
    that were parsed from these subtitles using the process described
    in Sec.~\ref{subsec:sentence-extraction}.
    \textit{Text Vocabulary} indicates the vocabulary across the sentences
    after removing punctuation, special characters, digits etc.
    \textit{Out-of-vocab} denotes the number of words that are not present
    in the training split,
    while \textit{Singletons} denotes the number of words appearing only once
    in the given partition.
    \textit{Duration} indicates the duration of the episodes.
    }
        \begin{tabular}{lccc\wipTwo{ccc}cccc}
            \toprule
            Split  & Episodes  &  Num.   & Num.          & Num.      & Sentence & Text & \wipTwo{Out-of-vocab & Singletons &} Avg. Duration & Total Duration \\
                   &           & Signers & Raw Subtitles & Sentences & Word Count & Vocabulary \wipTwo{& (O-O-V) & } & (mins) & (hours) \\
            \midrule
            train        & 1,675 & 28 & 1,108K & 1,004K &  9,557K & 72K & - & 22.0K & 44.3 & 1,236 \\
            val          & 33    & 7  & 22K  & 20K  &  205K & 14K & 0.8K & 6.1K & 50.2 & 28    \\
            test         & 254   & 4  & 192K & 168K &  1,593K & 35K & 4.8K & 11.9K &  48.2 & 204   \\
            \midrule
            total        & 1,962 & \totalSigners &  1,322K & \totalSentencesPrecise & \totalSentenceWordsPrecise & \totalVocabSentences & - & 23.6K & 45.1 & \totalHoursPrecise \\
            \bottomrule
        \end{tabular}
    \label{tab:splits}
\end{table*}

\subsection{Comparison to existing datasets}
\label{subsec:prior}
\begin{table*}[t]
    \setlength{\tabcolsep}{4pt}
    \caption{\textbf{Summary statistics of sign language datasets.} Language, co-articulated vs. isolated signing, sign vocabulary size, total number of sign annotations, corresponding spoken language vocabulary (if provided by dataset), total number of spoken language words, number of sequences, number of signers, source of data and duration in hours for each dataset.
    $^\dagger$Denotes the statistics of the subset of annotations used for sign language recognition experiments on these datasets, but in practice larger vocabularies are annotated (see Sec.~\ref{subsec:splits-rec} for details of annotations on \datasetName).
    }
    \centering
     \begin{tabular}{lcccccccccc}
      \toprule
      
      Dataset                                      & lang & co-articulated & sign & \#sign annots & text & \#words & \#sequences & \#signers & source    & \#hours  \\
                                          &  &  & vocab & (avg. per sign) & vocab &  & & &  &\\
      \midrule
     
      Devisign~\cite{chai2014devisign}             & CSL   & \xmark  & 2,000 & 24K (12) & - & - & -  & 8   & lab   & 13-33        \\ 
      CSL500~\cite{csl500} & CSL& \xmark & 500 & 125K (250) & - & - & -  & 50 & lab &  69-139\\
      ASLLVD~\cite{asllvid2008}                    & ASL   & \xmark  & 2,742 & 9K (3)  & - & - & -  & 6   & lab   &  4      \\
      ASL-LEX 2.0~\cite{sehyr2021asl}                  & ASL   & \xmark  & 2,723 &  2723 (1)   & - & - & -  & -   & lexicons, lab, web & - \\
      MSASL~\cite{Joze19msasl}                     & ASL   & \xmark  & 1,000 & 25K (25) & - & - & -  & 222 & lexicons, web & 25\\
      WLASL~\cite{Li19wlasl}                       & ASL   & \xmark  & 2,000 & 21K (11) & - & - & -  & 119 & lexicons, web & 14\\
      BSLDict~\cite{Momeni20b}   & BSL   & \xmark  & 9,283 & 14K (1) & - & -& -  & 148 & lexicons & 9 \\
      BosphorusSign22k~\cite{ozdemir2020bosphorussign22k} & TSL & \xmark  & 744 & 23K (30)& - & - & -   &6  & lab & 19  \\
      AUTSL~\cite{Sincan2020AUTSLAL} & TSL & \xmark & 226 &  38K (170)& - & -& -  &43  & lab  &  21 \\
      INCLUDE~\cite{Sridhar2020INCLUDEAL}& ISL& \xmark  &263 &4K (16)& - & - & -  & 7&lab & 3\\
      SMILE~\cite{smile} &DSGS &  \xmark & 100 &9K (90)& - & - & -  & 30 & lab & - \\
      \midrule
      S-pot~\cite{viitaniemi14}                    & FinSL & \cmark & 1,211 & 6K (5)   & - & - & 4K    & 5   & lab & 9\\ 
      Purdue RVL-SLLL~\cite{purdue06}              & ASL   & \cmark & 104  & 2K (19)  & 130 & 213 & -    & 14  & lab & -\\
      BOSTON104~\cite{EfficientApproxJointTrackingRecognition} & ASL &\cmark & 104 & 1K (10) & - & -& 201 & 3& lab &1 \\
      How2Sign~\cite{Duarte_CVPR2021} & ASL   & \cmark & - & - & 16K &  598K  & 35K &  11 &  lab & 79 \\ 
      CSL100~\cite{Huang2018VideobasedSL} & CSL   & \cmark & -  & -  & 178 & 175K & 25K  & 50  & lab & 100\\
      CSL-Daily~\cite{Zhou2021ImprovingSL} & CSL &\cmark & 2,000& 151K (76) & 2K & 312K & 21K & 10 & lab & 23 \\
      SIGNUM~\cite{signum2008}                     & DGS   & \cmark & 450  & 137K (304) & 1K & 166K & 33K   & 25  & lab & 	55\\
      Phoenix14T~\cite{Koller15cslr,Camgoz18}    & DGS   & \cmark & 1,066 & 76K (71)  &3K & 114K & 8K   & 9   & TV & 11 \\
      KETI~\cite{ko2019neural} & KSL  & \cmark & 524  & 15K (28) & - & - & - & 14  & lab  & 28 \\ 
      ISL~\cite{Kapoor2021TowardsAS}  & ISL   & \cmark & - & -  & 10K & - & 9K &  5 & web  & 18 \\
      GSL~\cite{adaloglou2020comprehensive}& GSL&\cmark & 310 & 41K & 481 & 44K & 10K  & 7 &lab & 10\\
      SWISSTXT-WEATHER~\cite{camgoz2021content4all} & DSGS & \cmark  & - & -  & 1K   & 7K  & 1K& -  & TV & 1\\ 
      SWISSTXT-NEWS~\cite{camgoz2021content4all} & DSGS & \cmark & -   & -  & 11K   & 73K & 6K& -& TV  & 9 \\
      SWISSTXT-RAW-WEATHER~\cite{camgoz2021content4all} & DSGS & \cmark & - & -  & -  & - & - & -  & TV  &  12 \\ 
      SWISSTXT-RAW-NEWS~\cite{camgoz2021content4all} & DSGS & \cmark & - & -  & -   & - & - & -  & TV  & 76 \\ 
      VRT-NEWS~\cite{camgoz2021content4all} & VGT & \cmark & - & -  & 7K   &80K& 7K & -  & TV  & 9\\ 
      VRT-RAW~\cite{camgoz2021content4all} & VGT & \cmark & - & -  & -   & - & - & -  & TV  & 100 \\ 
      BSL Corpus~\cite{schembri2013building}       & BSL   & \cmark & 5K   & 72K (14)  & - & - & -   & 249 & lab & 125\\ 
      BSL-1K~\cite{Albanie2020bsl1k}               & BSL   & \cmark & 1,064$^\dagger$ & 273K$^\dagger$ (257)  & 59K & 9M & 1M & 40  & TV  & 1,060 \\
      \midrule
      \textbf{\datasetName}                        & BSL  & \cmark & \vocabTwoKNoSpace$^\dagger$ & \numAnnosTwoKNoSpace$^\dagger$ (\annosPerSignTwoK) & \totalVocabSentences & 11.4M & 1.2M & 39  & TV & 1,467  \\
      \bottomrule \\
  \end{tabular}
\label{tab:prevdatasets}
\end{table*}

In Tab.~\ref{tab:prevdatasets}, we present a number of existing datasets used for sign language research -- mainly for the tasks of sign recognition, sign spotting, continuous sign language recognition, sign language translation and sign language production. Benchmarks have been proposed for American~\cite{asllvid2008,Joze19msasl,Li19wlasl,purdue06,EfficientApproxJointTrackingRecognition,Duarte_CVPR2021},  German~\cite{signum2008,Koller15cslr}, Swiss-German~\cite{smile,camgoz2021content4all}, Flemish~\cite{camgoz2021content4all}, Chinese~\cite{chai2014devisign, csl500, Huang2018VideobasedSL, Zhou2021ImprovingSL}, Finnish~\cite{viitaniemi14},  Indian~\cite{Sridhar2020INCLUDEAL,Kapoor2021TowardsAS},
Greek~\cite{adaloglou2020comprehensive}, Turkish~\cite{ozdemir2020bosphorussign22k,Sincan2020AUTSLAL}, Korean~\cite{ko2019neural} and British~\cite{Momeni20b,schembri2013building,Albanie2020bsl1k} sign languages.
These datasets can be grouped into \textit{isolated} signing
(where the signer performs a single sign, usually at a slow speed for clarity, starting from and ending in a neutral pose) and \textit{co-articulated} signing. Co-articulated signing,
or ``signs in context'', describes signing that exhibits variation
in sign form caused by immediately preceding or following signs,
or signs articulated at the same time.
If we are to build robust models which can understand sign language
\textit{``in the wild''}, we need to recognise co-articulated signs.

Most datasets in Tab.~\ref{tab:prevdatasets} fall into one or more of the following categories:
(i) They have a limited number of signers -- for example, Devisign~\cite{chai2014devisign}, ASLLVD~\cite{asllvid2008}, ISL~\cite{Kapoor2021TowardsAS}, GSL~\cite{adaloglou2020comprehensive} have 8 or fewer signers. (ii) They have a limited vocabulary of signs -- for example, Purdue RVL-SLLL~\cite{purdue06}, BOSTON104~\cite{EfficientApproxJointTrackingRecognition}, INCLUDE~\cite{Sridhar2020INCLUDEAL}, AUTSL~\cite{Sincan2020AUTSLAL}, SMILE~\cite{smile} only have a few hundred signs.
(iii) They have a large vocabulary of signs but only of isolated signs -- for example MSASL~\cite{Joze19msasl} and WLASL~\cite{Li19wlasl} have vocabularies of 1K and 2K signs, respectively.
(iv) They are recorded in lab settings.
(v) They are limited in total duration -- for example 
the popular PHOENIX14T~\cite{Koller15cslr} dataset contains 
only 11 hours of content.
(vi) They represent natural co-articulated signs but cover a limited domain of discourse -- for example, the videos in PHOENIX14T~\cite{Koller15cslr} and SWISSTXT-WEATHER~\cite{camgoz2021content4all} are only from weather broadcasts. 

BOBSL is most similar in content to PHOENIX14T~\cite{Koller15cslr},
SWISSTXT-WEATHER~\cite{camgoz2021content4all},
SWISSTXT-NEWS~\cite{camgoz2021content4all},
VRT-NEWS~\cite{camgoz2021content4all}
and BSL-1K~\cite{Albanie2020bsl1k}.
These datasets are all built from sign
language interpreted TV broadcasts.
PHOENIX14T~\cite{Koller15cslr},
SWISSTXT-WEATHER~\cite{camgoz2021content4all},
SWISSTXT-NEWS~\cite{camgoz2021content4all} and
VRT-NEWS~\cite{camgoz2021content4all} all provide 
valuable aligned subtitle annotations,
but are comparatively small in scale
(the latter three datasets also provide larger ``RAW'' unaligned variants akin to \datasetName
that are approximately an order of magnitude smaller than \datasetName in duration).
They are also restricted to a single domain of discourse:
weather broadcasts for PHOENIX14T~\cite{Koller15cslr}
and SWISSTXT-WEATHER~\cite{camgoz2021content4all};
news broadcasts for SWISSTXT-NEWS~\cite{camgoz2021content4all} and
VRT-NEWS~\cite{camgoz2021content4all}.
In contrast, BOBSL covers a variety of genres
(see Fig.~\ref{fig:pie_charts_genre}) and topics
(see Fig.~\ref{fig:pie_charts_topic}).
The relationship of BOBSL to the BSL-1K dataset is discussed in Sec.~\ref{subsec:relationship-to-bslonek}.

In summary, the BOBSL dataset presents several advantages:
it consists of co-articulated signs as opposed to isolated signs,
representing more natural signing
(note that BOBSL nevertheless remains distinct from conversational signing,
due to its use of interpreted content).
BOBSL provides the largest source of continuous signing (1,467 hours); it covers a large domain of discourse;
it is automatically annotated for a large vocabulary of more than 2,000 signs.
We note that since the annotations provided on the training and validation sets are obtained through automatic methods, they may contain some noise.
  
\subsection{Research use and potential changes}\label{subsec:usage}

BSL translation services are currently supplied to the BBC by Red Bee Media Ltd. They have indicated that they and their staff are happy for their footage to be used for research purposes. However, if the position changes the dataset will need to be revised accordingly. Researchers should be mindful of this, and should be aware that the ‘Permission to Use’ form they will need to sign obligates them to delete portions (or, indeed, the whole) of the dataset in the future, if so instructed.

\subsection{Relationship to the BSL-1K dataset}\label{subsec:relationship-to-bslonek}
In previous work~\cite{Albanie2020bsl1k},
we introduced the BSL-1K dataset, a collection of BSL videos
that were automatically annotated with sign instances via a keyword spotting method. 
This collection of automatic sign instances was further expanded 
through other methods for sign localisation \cite{Momeni20b, Varol21}.
A short test sequence was manually annotated for temporal sign segmentation
evaluation in~\cite{renz2021signtcn,renz2021sign}.
Manual signing sequence to corresponding subtitle alignments
have also been performed on BSL-1K in more recent work~\cite{Bull21}. 
However, BSL-1K remained as an internal dataset.
\datasetName represents a public, extended dataset based
on BSL-1K using
videos drawn from
the same source distribution
with no overlap between episodes to BSL-1K,
but significant overlap between signers and shows,
and preserving the same signer independent train, validation
and test split identities for signers that appear in both datasets.
The \datasetName dataset is larger
than \bslonek (1,467 hours vs 1,060 hours).
We have  automatically annotated BOBSL with sign instance timings using the same techniques as for BSL-1K and also provide signing sequence to corresponding text alignments. Through a data-sharing agreement with the BBC,
\datasetName is available for non-commercial research usage.

\section{\datasetName Dataset Construction}\label{sec:dataset-collection}

In this section, we describe the construction of
the \datasetName dataset.
We first describe the raw source data and
the pre-processing pipeline employed to prepare the data for
sign language research (Sec.~\ref{subsec:preproc}).
Next, we describe how the data was divided into
train, validation and test splits
(Sec.~\ref{subsec:partitioning})
and the automatic
methods
used to annotate this data with sign instance timings~(Sec.~\ref{subsec:auto-annotation}).
We detail the manual annotation processes we employ
(Sec.~\ref{subsec:manual-annotation})
together with our approach to subtitle sentence extraction
(Sec.~\ref{subsec:sentence-extraction}). Finally, we describe the BOBSL partitions for sign recognition (Sec.~\ref{subsec:splits-rec}), as well as for translation and alignment tasks (Sec.~\ref{subsec:splits-trans-align}).

\noindent \textbf{Dataset genesis.}
This dataset has been created in partnership with the British
Broadcasting Corporation (BBC), the UK's largest public service
broadcaster.  The UK broadcast regulator has set a
threshold for
the amount of accessible content broadcasters must supply. As a
result, the BBC produces subtitles for 100\% of its TV output, audio
description for more than 20\% of its output and BSL translations for
more than 5\% of its output. Due to the size of its weekly broadcast
output and its long-term retention of this metadata it has a
comparatively large datastore of useful data for partner universities
to work with.

The sort of data release represented by BOBSL is a core part of BBC
R\&D's remit as mandated by the UK 
Parliament\footnote{The 2016 Agreement with
the Department for Media, Culture and Sport mandates the BBC to
``ensure it conducts research and development activities geared
to ...maintain[ing] the BBC's leading role in research and development in
broadcasting ...in co-operation with suitable partners, such as
university departments ...'' (Section 65 of \url{http://downloads.bbc.co.uk/bbctrust/assets/files/pdf/about/how_we_govern/2016/agreement.pdf}).}.
As a result the BBC is keen to support research into accessibility
services by supplying data to partner universities and administering
non-commercial testing and training data to the wider academic
community.

\subsection{Source data and pre-processing}
\label{subsec:preproc}

\noindent \textbf{Source data.}
An initial collection of TV episodes were provided by the BBC.
These were broadcast between 2007 and 2020
and vary from a few minutes to 120 minutes in duration
(see Fig.~\ref{fig:episode-durations} for the distribution of episode durations).
Each episode is accompanied by a corresponding set of written
English subtitles, derived from the audio track of the show.
The programs span a wide range of topics (history, drama,
science etc.)---a detailed summary of the content included
is provided in Sec.~\ref{subsec:content}.
The majority of these shows are accompanied by a BSL interpreter,
overlaid on the bottom right hand corner of the screen
in a fixed location.
Note that sign interpreters produce a \textit{translation} of the speech
that appears in the subtitles, as opposed to a \textit{transcription}.
This means that words in the subtitles may not correspond directly to
individual signs produced by the interpreters, and vice versa.
The videos have a height of 576x pixels,
a display aspect ratio of 16:9 and a frame rate of 25 fps.

\vspace{4pt}

\noindent \textbf{Filtering and pre-processing.}
First, TV programs that were known to not contain a BSL interpreter
in a fixed region of the screen were removed from the collection.
A small number of videos that exhibited significant data corruptions
were also removed.

\begin{figure}[ht]
    \centering
    \includegraphics[trim={0 16cm 31cm 0},clip,width=0.48\textwidth]{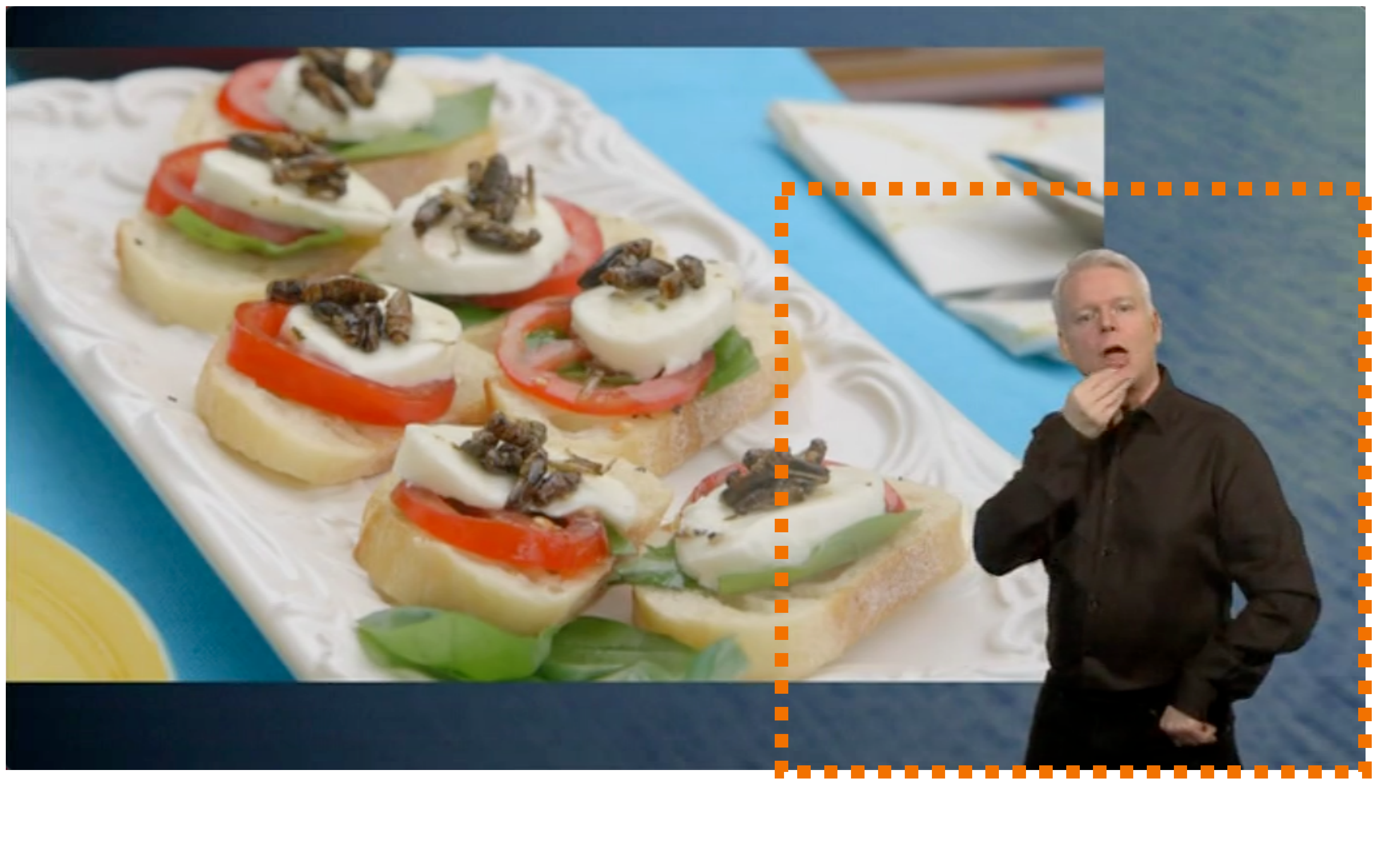}
    \caption{
    \textbf{Pre-processing.}
    Raw broadcast footage is pre-processed  by extracting a
    $444\times444$ pixel square crop from the bottom right-hand corner
    region occupied by the BSL interpreter in each video
    (illustrated by the orange dashed box).
    }
    \label{fig:pre-processing}
\end{figure}

\begin{figure}
    \centering
    \includegraphics[width=0.48\textwidth]{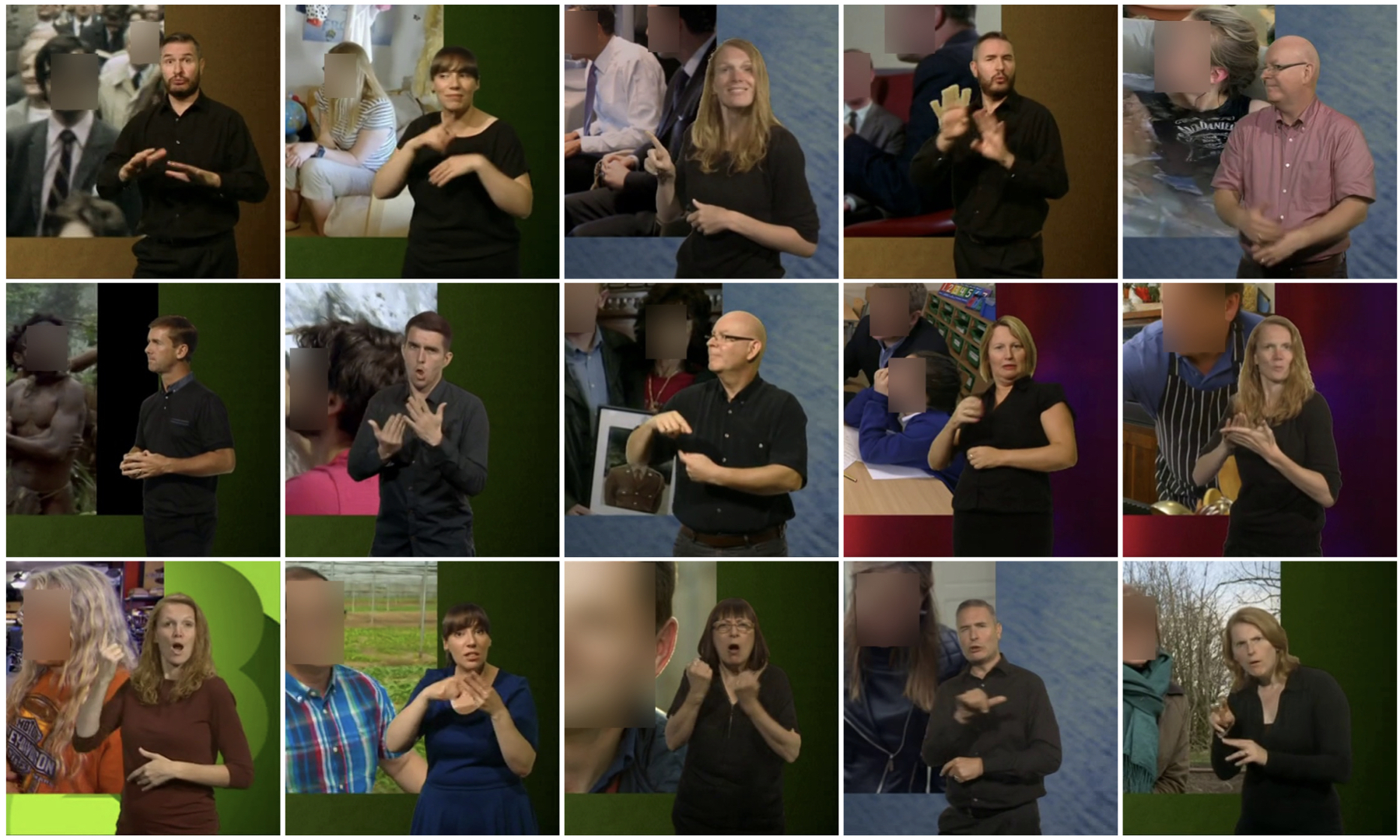}
    \caption{
    \textbf{Background face blurring.} Faces appearing behind the interpreter
    are automatically tracked and blurred for anonymisation purposes.
    }
    \label{fig:blur}
\end{figure}

\noindent \textit{Video pre-processing.}
Each video was cropped to include only the bottom-right
$444\times 444$ pixel region containing the BSL interpreter
(see Fig.~\ref{fig:pre-processing}).
We employed the automatic face detection and tracking pipeline
provided by the authors of~\cite{Chung16a} to detect and track faces, 
with the goal of blurring those appearing in the content behind the intepreter.
For this it was first necessary to determine which
face tracks belong to the interpreters and exclude them. 
To that end we extracted pose estimates from each frame
using OpenPose~\cite{cao2019openpose}
and employed a heuristic to determine background face tracks
(tracks with duration shorter than 5 seconds or that do not
exhibit overlap with the estimated keypoints of the interpreter).
The pixels under all the background face tracks are blurred using a Gaussian filter.
Using this pipeline we blur 224,957 face tracks over 170 hours of video.
Some examples are shown in Fig.~\ref{fig:blur}.
We observe qualitatively that the pipeline performs well for clearly visible background faces.
However, we note a limitation of our approach: the automatic face detector
can make mistakes (typically for cases in which the background face is very small or heavily occluded)
and thus there are likely to be a small number of background faces that are not blurred.

\noindent \textit{Subtitle pre-processing.}
After manual inspection, we observed that approximately one quarter of the subtitle 
files exhibited discrepancies in time alignment between the audio track and the subtitle timestamps.
To address these cases,
we applied standard methods of forced alignment using an acoustic model\footnote{\url{https://www.readbeyond.it/aeneas/}, \url{https://subsync.online/}}.

After pre-processing the videos and subtitles,
the audio track of each video was removed. 
The final result of these filtering and pre-processing steps was a
collection of 1,962 videos containing BSL interpreters with
corresponding audio-aligned written English subtitles that
form the public dataset release.

\subsection{Dataset splits}
\label{subsec:partitioning}

To support the development of signer-independent systems
(in which models are evaluated on signers not seen during training),
we divide the dataset into train, validation, test splits according
to the estimated identity of the BSL interpreters.

To determine the interpreter identity associated with each video, we
employ a semi-automatic process.
We first detect the face of the interpreter in a 10-second
clip extracted from the temporal midpoint of the video
(since the interpreter does not change over the course of
a single program,
a short clip suffices to perform identification and
reduces computational cost relative to using the full program).
This is done with a RetinaFace face detector~\cite{deng2019retinaface}
that employs a MobileNet0.25 trunk architecture~\cite{howard2017mobilenets}.
The model is trained for face detection on the WIDER face
benchmark~\cite{yang2016wider}.
Next, face embeddings are computed from each detected face
bounding box with an SE-50~\cite{hu2018squeeze} face verification network.
The face detections are then linked into tracks by minimising
a cost function based on spatial overlap between face detections
and similarities between face embeddings.
For each track, the embeddings from the face detections are aggregated
via averaging and then L2-normalised to produce a single track descriptor.
Next, agglomerative clustering is used to group the track descriptors into
an initial set of identity clusters
(we employ the implementation provided by~\cite{scikit-learn},
with a distance threshold of 0.32 over the cosine similarities between descriptors).
Following this, the identity clusters are checked manually, with erroneous
assignments corrected by re-assigning to the correct cluster.
As a result of this process, 39 identities were identified.

Finally, signers are assigned to separate splits,
to produce the dataset statistics given in Tab.~\ref{tab:splits}.
The distribution of episodes associated to each signer,
together with the split information is illustrated in Fig.~\ref{fig:signer-dist}.

\subsection{Automatic annotation via sign spotting and localisation methods}
\label{subsec:auto-annotation}

\begin{figure}%
    \centering
    \includegraphics[width=0.495\textwidth]{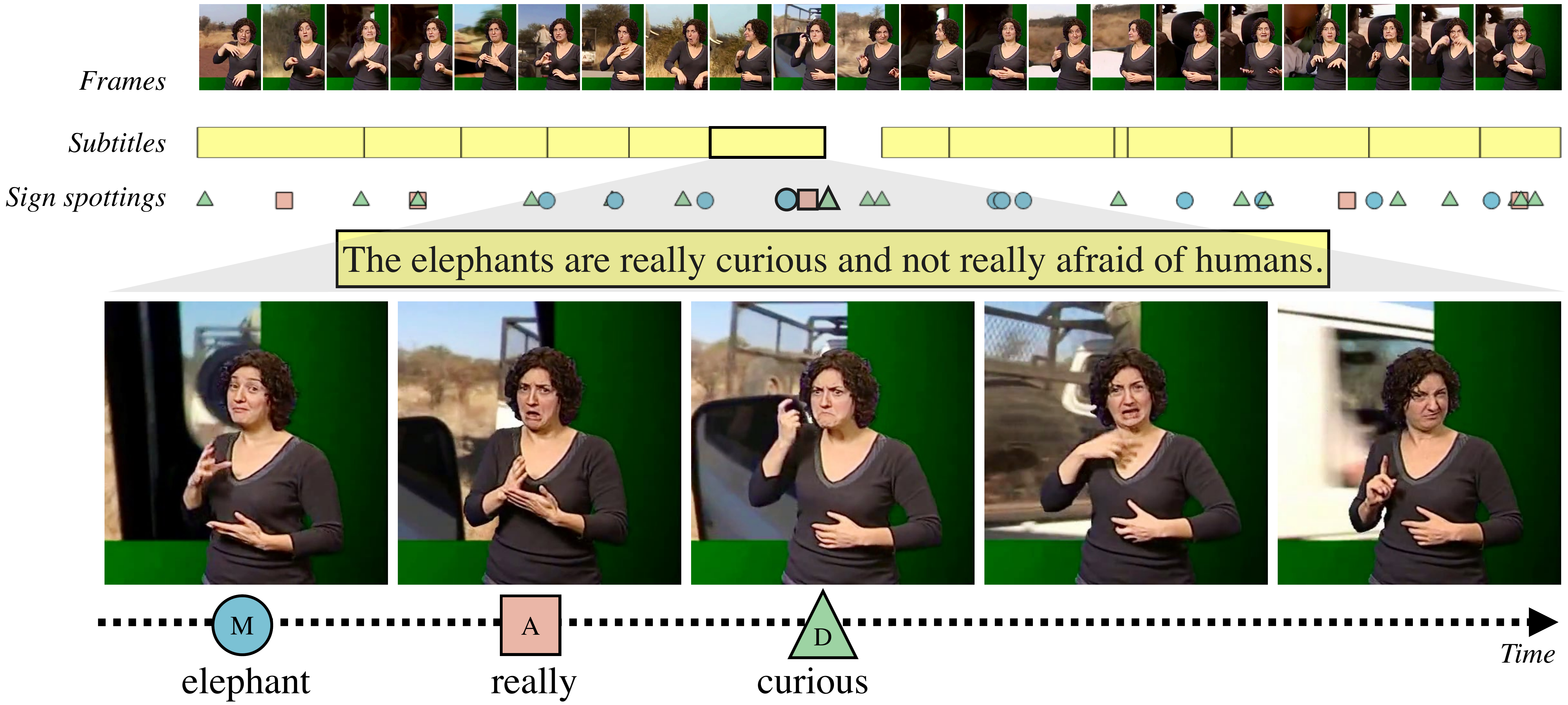}
    \caption{
        \textbf{BOBSL sample with automatic sign annotations.}
        We show a sample training video, together with the corresponding English language subtitle, and automatic annotations generated through three sign spotting techniques (\textit{M}: mouthing, \textit{D}: dictionary, \textit{A}: attention, described in Sec.~\ref{subsec:auto-annotation}).
    }
    \label{fig:teaser}
\end{figure}

Due to the large scale of the \datasetName dataset,
exhaustive manual annotation of individual signs would be prohibitively expensive.
We therefore turn to automatic annotation techniques for sign instance localisation
making use of the information within weakly-aligned subtitles.
In particular, we employ:
(1)~the mouthing keyword spotting approach from~\cite{Albanie2020bsl1k}, 
(2)~the dictionary spotting approach from~\cite{Momeni20b},
and (3)~the attention spotting approach from~\cite{Varol21}
to annotate the data.
We give a brief summary of each method here
and refer the reader to the original papers for further details.
Fig.~\ref{fig:teaser} provides sample annotations from each method
on a sample training video.

\vspace{4pt}

\noindent(1) \textbf{Keyword spotting with mouthings.}
A sign may consist of not just movements of the hands, but also
head movements, facial expressions and mouthings~\cite{sutton1999linguistics}. 
Mouthings have multiple roles:
they can be used to specify the meaning of a sign in the case of
polysemy and to disambiguate manual homonyms~\cite{woll2001u}.
Mouthings appear frequently in BSL - accompanying
over 2/3 of signs in one study~\cite{sutton2007mouthings}.
From an annotation perspective,
mouthings provide a cue for \textit{sign spotting},
the task of localising a given sign in a signing sequence.

In this work, we employ the method proposed in~\cite{Albanie2020bsl1k} to spot signs.
This method works in several stages:
First, given a target ``keyword'' for which we wish to spot signs,
we find all occurrences of the keyword in the subtitles. 
Next, for each subtitle containing the keyword, we pad its temporal extent by several
seconds to create a search window in which the sign has a high probability of occurring.
Through preliminary experiments, we found that padding by 10 seconds
on both sides worked well.
Finally, we employ a keyword spotting model to find whether and when the mouthing
occurs within the constructed search window. The model outputs a confidence score
associated to each frame, we record all localisations above 0.5 threshold as our
automatic mouthing annotations (after a non-maximum suppression stage as in~\cite{Albanie2020bsl1k}). Fig.~\ref{fig:recognition-train} provides
statistics for the amount of annotations on the training set.

To derive the list of candidate keywords for spotting,
we first apply \textit{text normalisation} to the subtitles
using the method of~\cite{flint2017text}.
This normalisation converts dates and numbers to their written form,
e.g.\ 13 becomes ``thirteen''.
From BOBSL subtitle words, we obtain 79K
(we use the original subtitles, rather than sentences
for spotting---the vocabulary differs slightly due to the filtering
involved in sentence extraction)
and 72K unique words before
    and after text normalisation, respectively.
We further filter the list of keywords to those that appear in the CMU phonetic dictionary~\cite{cmu} with at least four phonemes
(the model is trained on words with at least 6 phonemes, but we found 4 to work reasonably). This filtering results in a final list of 43K search keywords.

\begin{figure}
    \centering
    \includegraphics[width=0.48\textwidth]{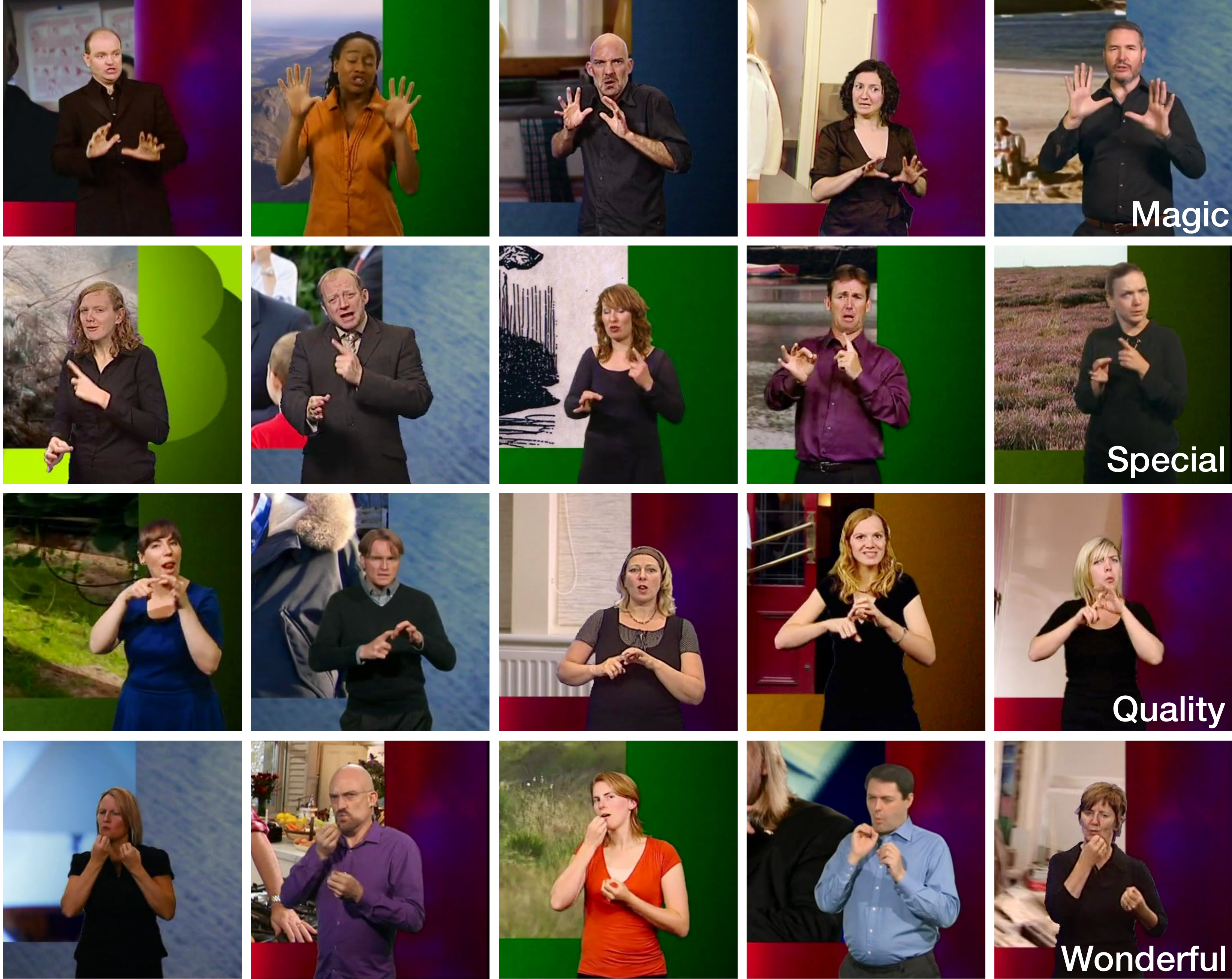}
    \caption{
    \textbf{BOBSL automatic sign annotations through mouthings.}
    We show examples of automatically retrieved instances of four different signs on each row (\textit{magic}, \textit{special}, \textit{quality}, \textit{wonderful}) obtained through the pipeline of keyword spotting with mouthings.
    }
    \label{fig:sign_samples}
\end{figure}

The keyword spotting model used is an improved variant of the model of Stafylakis~et~al.~\cite{stafylakis2018zero}
from \cite{Momeni2020kws} (described in their paper as 
``P2G \cite{stafylakis2018zero} baseline'').
The model is trained on ``talking heads'' datasets
(LRW~\cite{chung2016lip} and LRS2~\cite{chung2016signs})
of BBC TV broadcasts.
While the model has never been trained on signers,
we observe that it generalizes well to a large set of signer mouthings.
As observed in~\cite{Albanie2020bsl1k},
the peak in the posterior probability assigned to the presence of a keyword 
typically corresponds to (approximately) the end of the mouthing/sign.
Qualitative examples of automatically retrieved signs through this method
are shown in Fig.~\ref{fig:sign_samples}.

\begin{figure}
    \centering
    \includegraphics[width=0.48\textwidth]{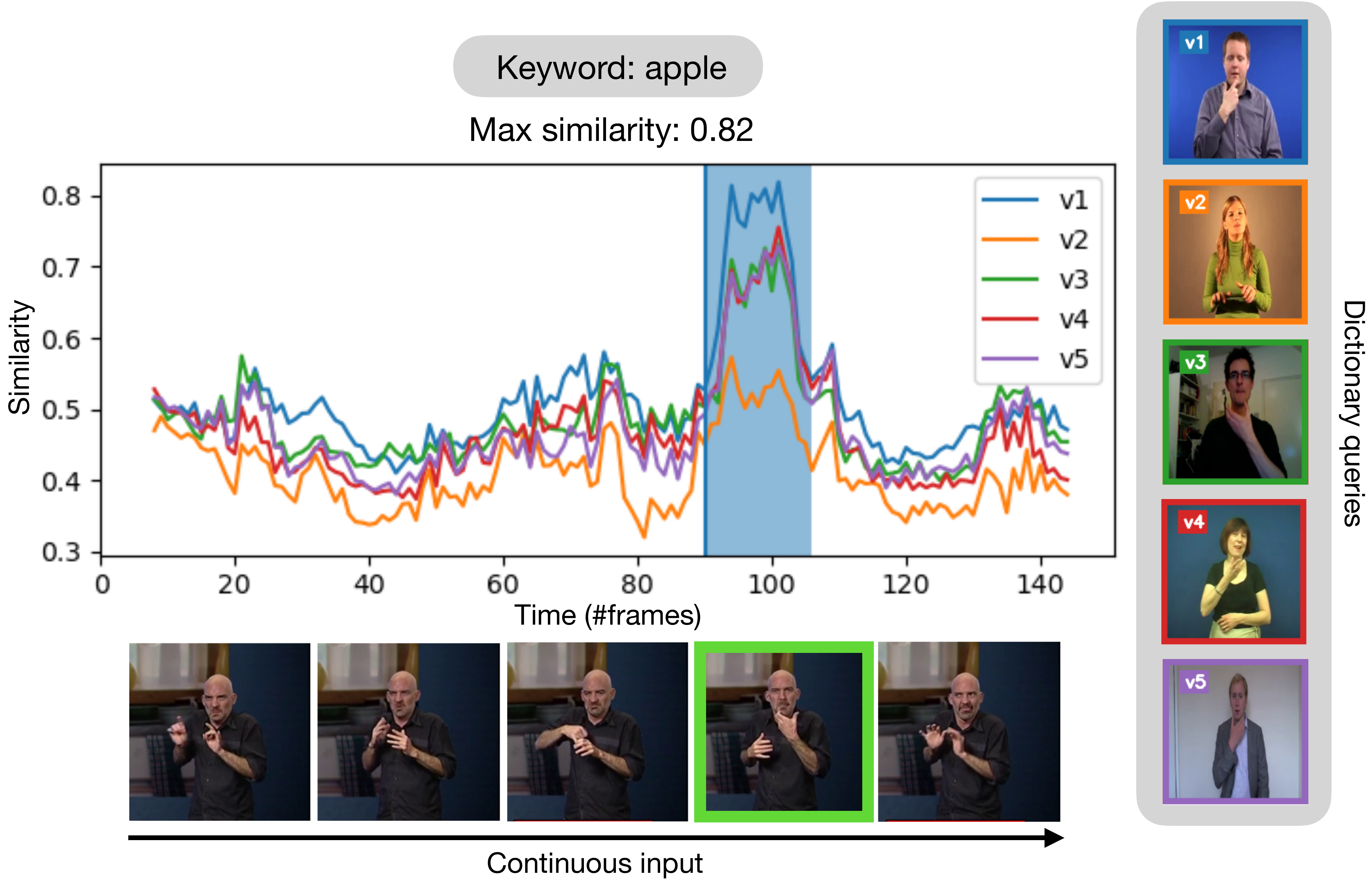}
    \caption{
        \textbf{BOBSL automatic sign annotations through dictionaries.}
        We illustrate the localisation procedure for comparing dictionary samples
        for a given keyword with a continuous signing.
    }
    \label{fig:sign_samples_dict}
\end{figure}

\vspace{4pt}

\noindent(2) \textbf{Sign spotting with dictionaries.} 
Following the method proposed in \cite{Momeni20b},
given a video of an isolated sign from a dictionary,
we identify whether and where it has been signed
in a continuous, co-articulated sign language video.
To this end, we learn a joint embedding space where we can
measure similarity between isolated dictionary videos
and continuous signing.
This method leverages the weakly-aligned subtitles by querying words
in the subtitle within a $\pm$4sec padded neighbourhood around the subtitle timestamps
(note in practice we use sentences instead of subtitles, see Sec.~\ref{subsec:sentence-extraction}).
In particular, we query words and phrases from the \bsldict~\cite{Momeni20b}
vocabulary if they occur in the sentences.
In order to determine whether a query from the
dictionary occurs in the sentence,
we check the sentence in its original,
and lemmatised forms and the query in its
original and text-normalised forms.
If a match is found, we query the dictionary video(s)
corresponding to the word/phrase.

In order to obtain the embedding space, we follow a slightly different procedure
than \cite{Momeni20b} for simplicity. We only perform the first stage of
\cite{Momeni20b}, which is to train an I3D classification
model jointly on continuous annotations and \bsldict samples.
We do not further train the MLP network on top of the I3D
features with contrastive loss. Instead, we initialise the weights
of the I3D with a stronger recognition model provided by \cite{varol21bslattend},
trained for 5K categories from the mouthing and dictionary annotations of BSLK-1K~\cite{Albanie2020bsl1k}.
We do not re-initialise the batch normalisation layers unlike \cite{Momeni20b}).
For joint finetuning, we use the mouthing (threshold=0.8) and
dictionary (threshold=0.8) spottings from BSL-1K, as well as \bsldict
videos filtered to the 1K vocabulary of \cite{Albanie2020bsl1k}.
We found the features from this model to be sufficiently strong
to provide us automatic sign annotations on \datasetName.

We obtain a single embedding for the dictionary sample
by averaging features computed with multiple frame rates as in \cite{Momeni20b}.
We obtain a sequence of embeddings for the
\datasetName search window by applying a sliding window with
a stride of 4 frames.
We compute the similarity between the continuous signing
search window and each of the dictionary variants for a
given word/phrase: we record the location where the similarity
is maximised for all variants and choose the best match
as the one with highest similarity score.
We record all localisations above 0.7 threshold as our automatic dictionary
annotations. Fig.~\ref{fig:recognition-train} provides
statistics for the number of annotations on the training set.
We refer to Fig.~\ref{fig:sign_samples_dict}
for an illustration of the similarity plots across variants.

\vspace{4pt}

\noindent(3) \textbf{Sign localisation with Transformer attention.}
In contrast to the two previous automatic annotation methods,
the approach~\cite{Varol21} of localising signs differs considerably in that it is 
\textit{context-aware}. We train a Transformer model~\cite{Vaswani2017} to predict, given an 
input stream of continuous signing, the sequence of corresponding written tokens. We then perform
sign localisation by using the trained attention mechanism of the Transformer to align written 
English tokens to signs. More specifically, once the model is trained, new sign instances are 
localised for tokens that have been correctly predicted by determining the index at which the 
corresponding encoder-decoder attention is maximised. 
We observe that even low values for the maximum attention score
provide good localisations; therefore, we do not apply any threshold for attention
spottings.
Fig.~\ref{fig:recognition-train} provides
statistics for the amount of annotations on the training set.

In practice, we train the Transformer on a subset of video-text pairs, which contain at least one
sign automatic annotation (from the two previously described methods) within the sentence 
timestamps. In such a way, we ensure there is an approximate alignment between the source signing
video and target written token sequence. The encoder input video is represented by a 
1024-dimensional feature
sequence, extracted from an I3D model provided by \cite{varol21bslattend} which is trained on 
sign classification with BSLK-1K~\cite{Albanie2020bsl1k} for a 5K vocabulary of signs (obtained 
from mouthing and dictionary spottings) applied with a sliding window of stride 4.
For building the target written sequences, we (1)~lemmatise the words in every sentence assuming 
inflected versions of the same word map to the same sign, (2)~filter to a vocabulary of 18K 
lemmas obtained by combining the automatic annotations from mouthing (threshold=0.7) and 
dictionary (threshold=0.8) spottings, and (3)~remove stop words.

Recent work has also demonstrated the effectiveness of the
Transformer for sign spotting with
dictionaries~\cite{jiang2021looking}---we defer an investigation
of this approach to future work.

\begin{figure}[t] 
    \centering
    \includegraphics[width=.48\textwidth,trim={0cm 0cm 0cm 0cm},clip]{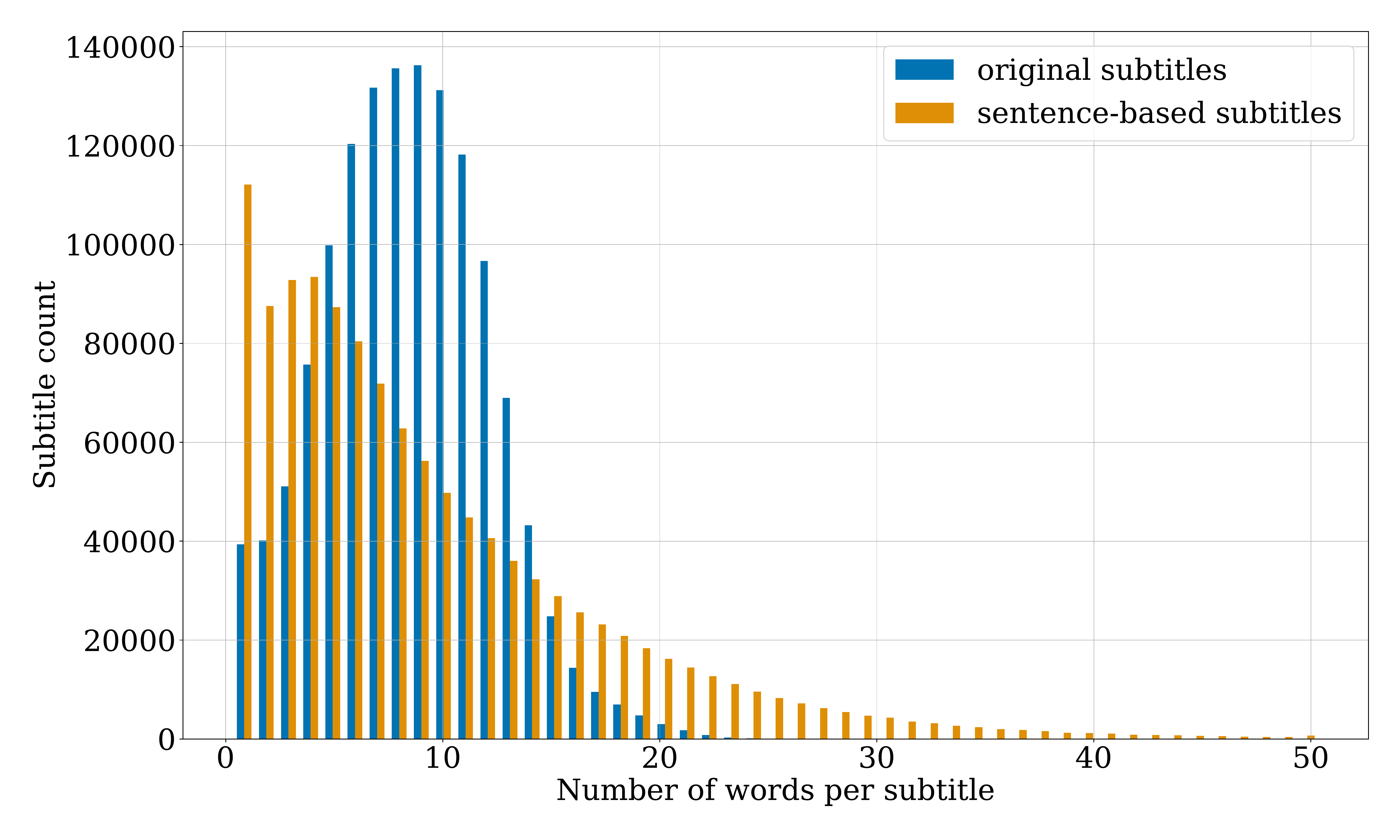}
    \caption{\textbf{The distribution of lengths for the original and sentence-aligned subtitles.}
    In contrast to the broadcast subtitles which possess a relatively small variance in length,
    the sentence-based subtitles exhibit a broader variance, with a greater number of very short (just a few words)
    and very long (more than 30 words) sequences.
   }
    \label{fig:sentence-distribution}
\end{figure}

\vspace{4pt}
\subsection{Sentence extraction}
\label{subsec:sentence-extraction}

The subtitles associated with the \datasetName episodes are approximately aligned to the audio track of the corresponding content
but do not necessarily fall into well-formed sentences. 
To support research into tasks such as sign language translation
(which often operates at the sentence-level~\cite{Camgoz18,camgoz2021content4all})
we extract well-formed sentences from the subtitles.
This is done semi-automatically by splitting subtitles on sentence boundary punctuation 
and employing a combination of heuristics and manual inspection to resolve ambiguous cases.
To preserve an approximate time alignment between the sentences and the signing,
when multiple sentences fall within a single subtitle, we employ a further simple heuristic:
each sentence is assigned a duration in proportion to its written length (in characters)
as a fraction of the original subtitle.
Finally, we remove sentences that correspond to descriptions of background music lyrics
(these are typically unsigned) and sentences that are known to fall outside the feasible
signing period (e.g.\ those that occur after the show credits).
The result of this sentence extraction process is a collection of ``sentence-based'' subtitles
(in which each subtitle corresponds to a single sentence), summarised in Tab.~\ref{tab:splits}.
In comparison to the original subtitles (which are relatively uniform in duration)
the distribution of sentence lengths exhibits broader variance
(this effect is visualised in Fig.~\ref{fig:sentence-distribution}).
Note that since the sentence extraction process makes use of punctuation
in the subtitles,
some long subtitles may be due to missing punctuation:
a manual inspection of random samples determined that
this occurs relatively rarely.

\begin{figure}[t]
    \centering
    \includegraphics[width=.48\textwidth,trim={0cm 0cm 0cm 0cm},clip]{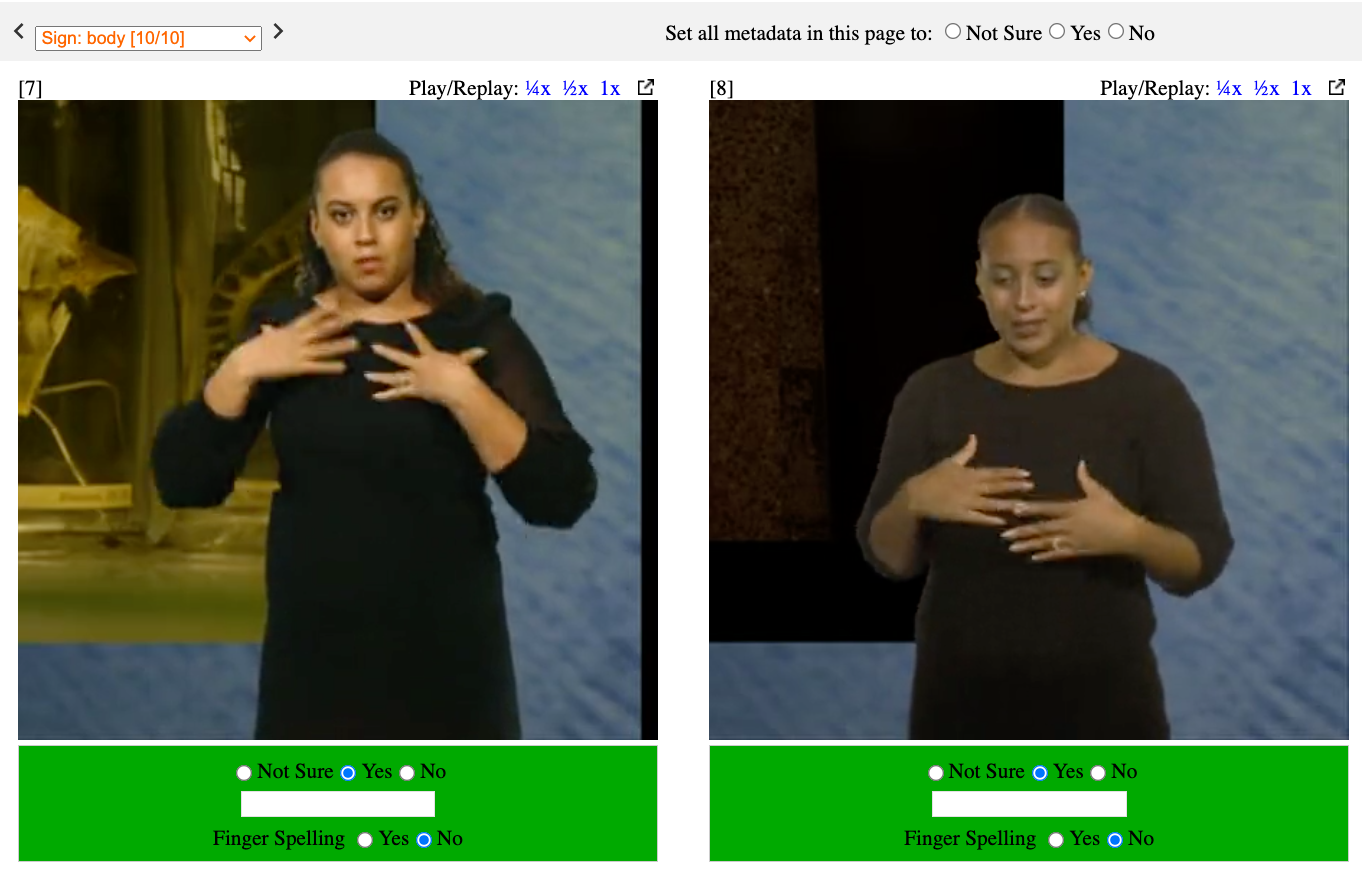}
    \caption{\textbf{Manual annotations.}
    A screenshot of the VIA Whole-Sign Verification Tool.
    Given proposed temporal windows from the automatic sign spotting
    methods described in Sec.~\ref{sec:dataset-collection},
    annotators can mark the proposals as correct, incorrect or unsure,
    and provide additional metadata (see Sec.~\ref{subsec:manual-annotation} for details).
     }
    \label{fig:via-screenshot}
\end{figure}

\subsection{Manual annotation}
\label{subsec:manual-annotation}

\noindent \textbf{Sign verification.}
Deaf annotators proficient in BSL used a variant of the
VIA tool that was adapted for whole-sign verification~\cite{dutta2019via}
(see Fig.~\ref{fig:via-screenshot}),
similarly to the process used by~\cite{Albanie2020bsl1k}.
To enable efficient collection,
labels were collected for temporal proposals for signs
in the test split by verifying/discarding automatic
spottings that were assigned high confidence scores
by
the automatic sign spotting techniques
(above 0.9 confidence for mouthing annotations,
above 0.8 for the dictionary annotations).
When viewing a temporal proposal, the video could be played
at different speeds (and replayed if needed).
For each proposed spotting location,
the annotator is able to indicate:
(i)~whether the sign is correct, incorrect, or that they are unsure,
(ii)~whether fingerspelling (using the manual alphabet to spell English words) was used,
(iii)~further comments, including the meaning of the sign
(if the proposed meaning was incorrect),
and any other observations.

For quality control, a small random sample of the annotations were
further verified by a deaf native signer of BSL.
Of the mouthing spottings within the \vocabTwoK vocabulary
(this vocabulary is described in more detail in Sec.~\ref{subsec:splits-rec})
with a confidence
of at least 0.9 that were annotated, 63.6\% were marked correct,
yielding 9,263 verified signs spanning 1,653 classes.
The latter figure includes predictions that were corrected
by annotators,
as well as a small number of verified low confidence signs
that were annotated during early development.
Of the dictionary spottings within the \vocabTwoK vocabulary with a confidence
of at least 0.8 that were annotated,
75.8\% were marked correct,
yielding 15,782 verified signs (spanning 765 classes)
after including corrections.
These verification statistics also exclude a small number signs that
were tagged by annotators as ``inappropriate'' in modern BSL signing.

\begin{figure}[t]
    \centering
    \includegraphics[width=.48\textwidth,trim={0cm 0cm 0cm 0cm},clip]{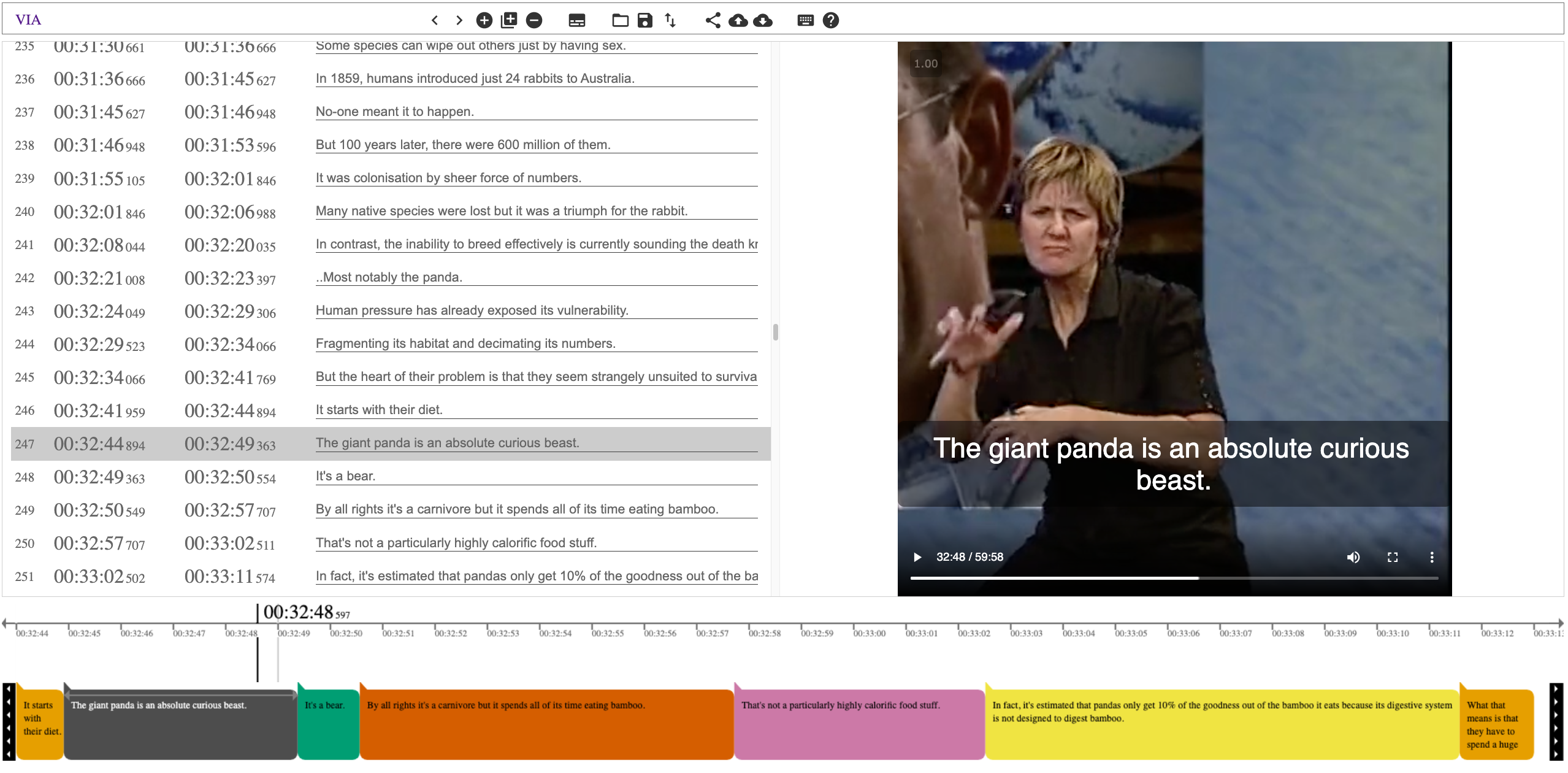}
    \caption{\textbf{Sentence alignment tool.}
    A screenshot of the VIA sentence alignment tool \cite{dutta2019via}.
    The annotator uses a ``draggable'' visualisation of the
    temporal extent of the sentences at the bottom of the screen to perform
    alignment, with the ability to pause and replay segments.
   }
    \label{fig:via-subtitle-alignment-screenshot}
\end{figure}

\begin{figure}[t]
    \centering
    \subfloat[]{\label{fig:distribution_start}\includegraphics[width=0.5\linewidth]{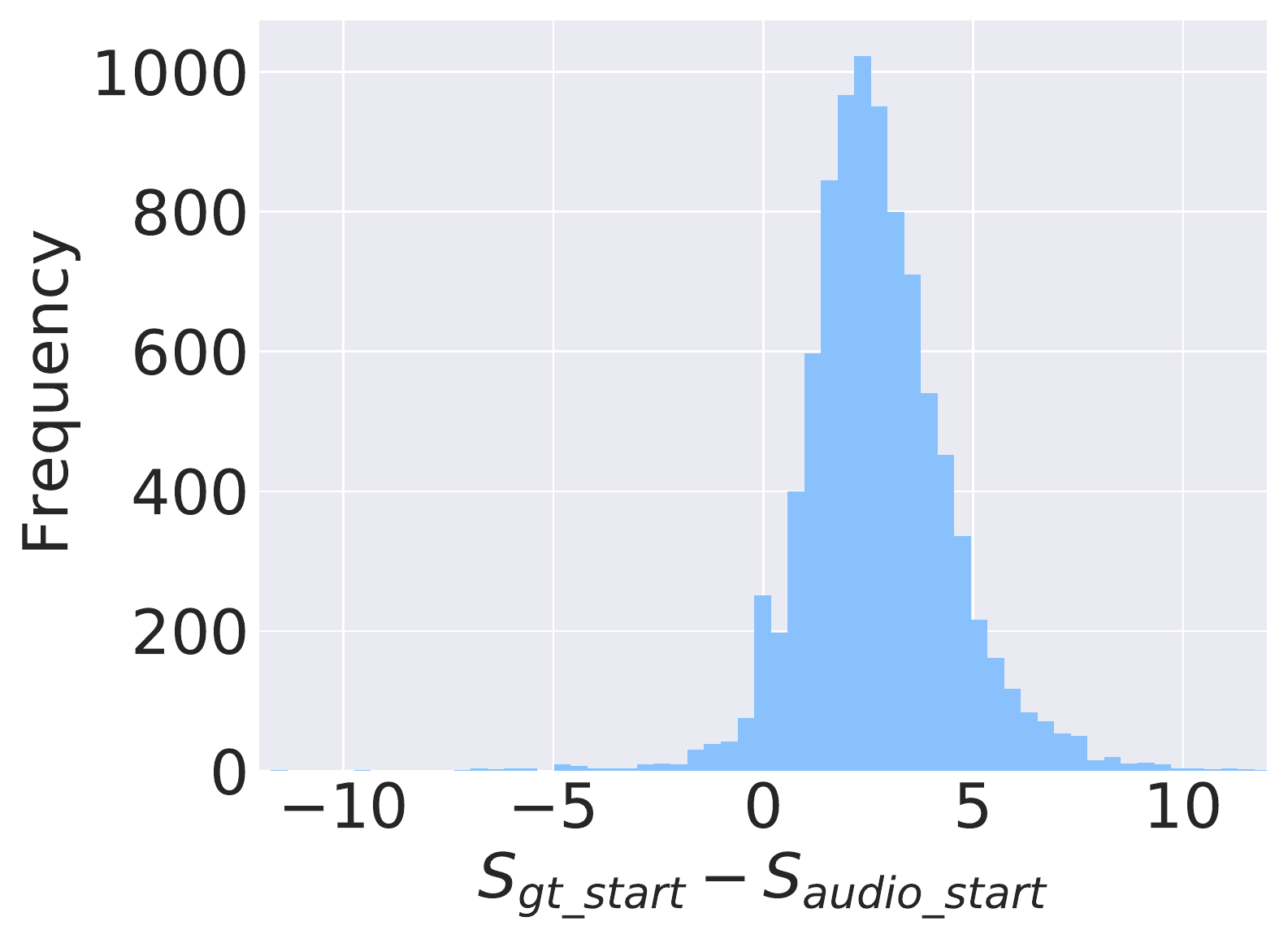}}\hfill
    \subfloat[]{\label{fig:distribution_dur}\includegraphics[width=0.5\linewidth]{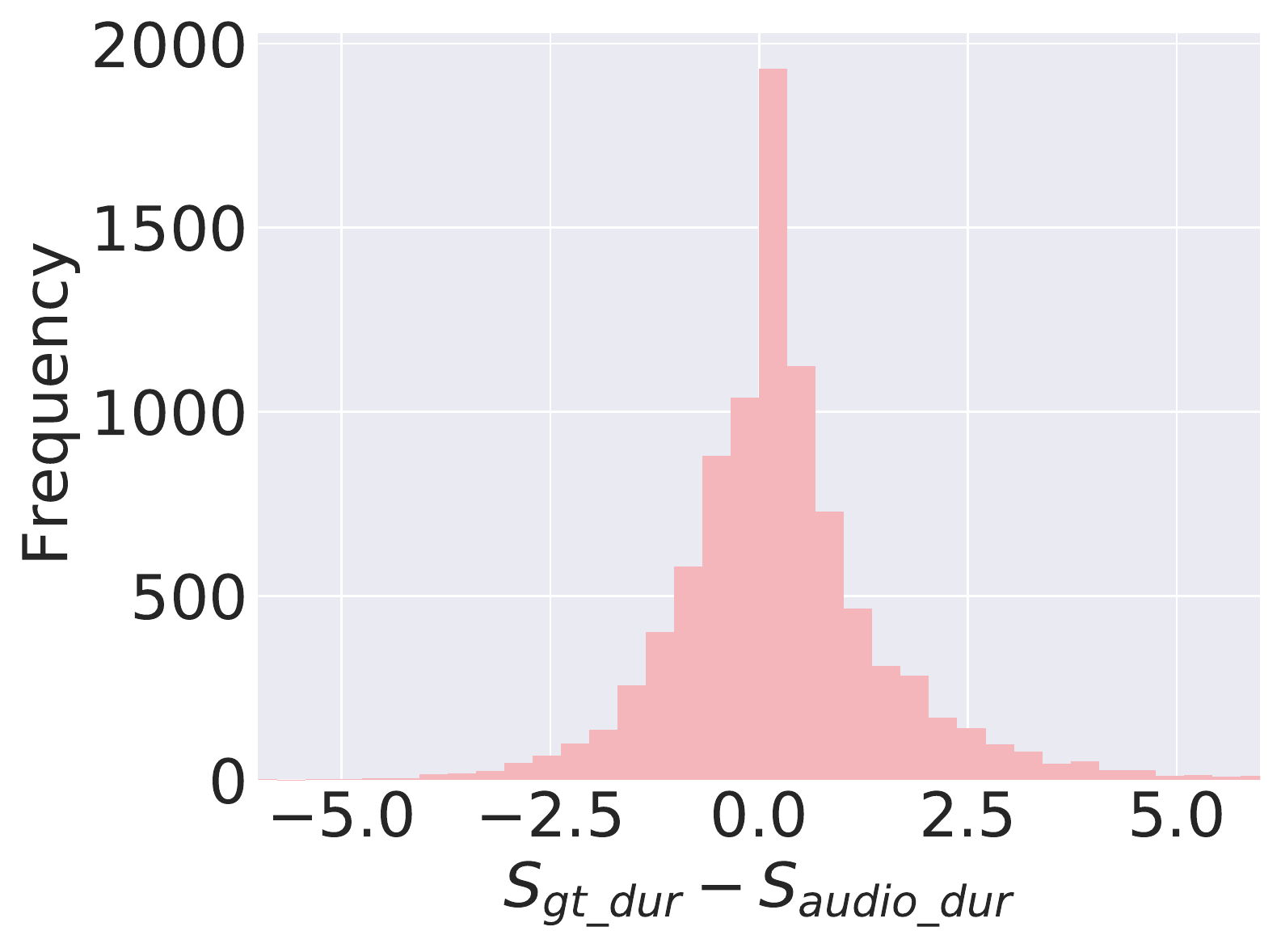}}
   \caption{\textbf{Audio-aligned versus signing-aligned subtitles.}
    We plot the distribution of temporal shifts between the signing-aligned and audio-aligned subtitles for \trainManual by showing the differences in subtitle (a) start times and (b) duration.
     }
    \label{fig:audio-signing-subtitle}
\end{figure}

\noindent \textbf{Sentence alignment.} 
To support research into the tasks of
sign language alignment and translation,
we manually align the extracted sentences
(which are initially coarsely aligned with the audio content)
with the signing content for a subset of the episodes. 
The audio-aligned sentences differ from the signing-aligned
subtitles in both start time and duration, as shown in Fig.~\ref{fig:audio-signing-subtitle}.
To perform the alignment,
we used an adapted version of the VIA tool,
shown in Fig.~\ref{fig:via-subtitle-alignment-screenshot}.
The annotator is presented with a list of sentences
for which they are able to adjust timings by
clicking and dragging elements on a webpage
(this methodology is similar to the alignment tool described
in the concurrent approach of~\cite{camgoz2021content4all}).
We make these sentence-level alignment annotations available.

\subsection{BOBSL partitions for \textit{sign} recognition evaluations}\label{subsec:splits-rec}

In order to 
evaluate the performance of sign recognition models, we provide (i) large \textit{automatic} training and validation sets of sign instances as well as a (ii) large \textit{human-verified} test set for benchmarking.
\vspace{4pt}

\begin{figure*}
    \centering
    \includegraphics[width=.99\textwidth]{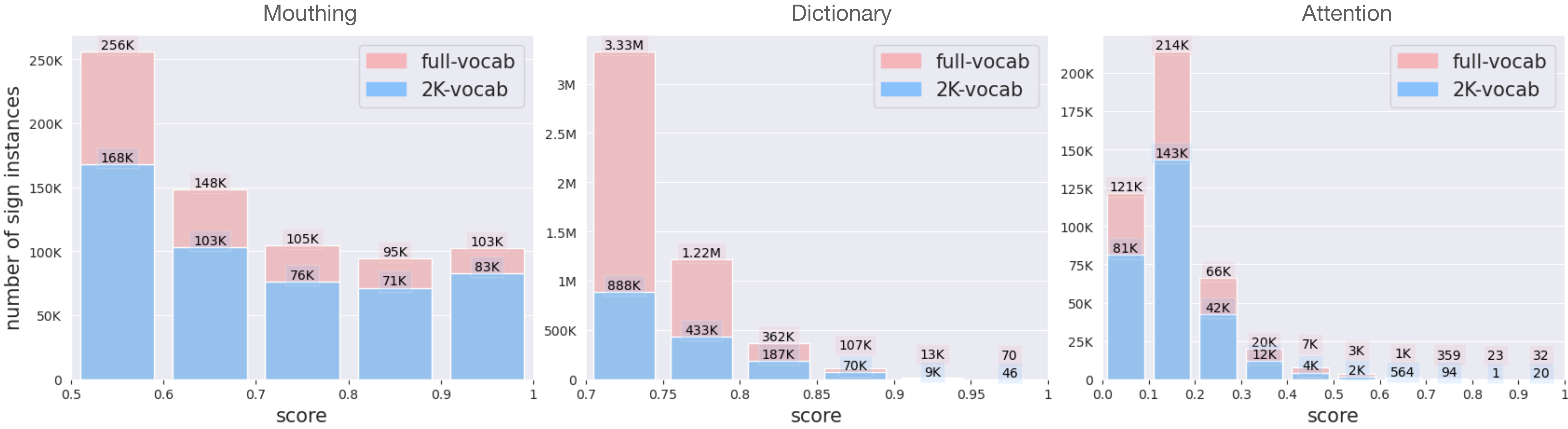}
    \caption{\textbf{Automatic training set of sign instances.}
        We obtain several million automatic annotations
        through sign spotting, with varying levels of noise.
        We show the
        different subsets of the training set obtained from mouthing~(M), dictionary~(D) and attention~(A) spottings according to their
        confidence scores. Note that the range of scores for each annotation
        type is different; we observe that minimal thresholds of
        0.5, 0.7, 0.0 are necessary for M, D, A, respectively. In practice,
        we train our recognition models on a subset
        of the annotations (M and D spottings 
        from the \vocabTwoK vocabulary that are above 0.8 score)
        to retain an affordable training time with high quality annotations.
    \label{fig:recognition-train}
    }
\end{figure*}

\noindent \textbf{Recognition vocabulary.} The construction of an appropriate vocabulary set for sign recognition
is a challenging linguistic task for several reasons.
First, BSL grammar differs significantly from English grammar
(for instance, while English typically adds an ``s'' suffix to indicate plurality, 
BSL has several ways of marking a noun plural~\cite{sutton1999linguistics},
e.g.\ through repetition, quantifier signs and whole-sign modification).
More broadly, there is a complex many-to-many mapping between English words and BSL signs,
and there are many signs that correspond to sequences of English words.
Additionally, in BSL, fingerspelling can be used to express English words that have no sign equivalents (for example, proper names).

In the absence of standard writing systems for sign languages~\cite{johnston2010archive},
a number of different gloss systems have been developed and used for corpus linguistics~\cite{johnston2010archive,schembri2013building}.
A central challenge in adopting such approaches is scalability:
providing fine-grained, consistent linguistic sign glosses requires 
a highly skilled team of annotators and vast labour investments
(for example, the BSL Corpus~\cite{schembri2013building}
is an extensive ongoing research effort spanning more than a decade---even so,
it has only been practical to label a relatively small fraction of the total
signing data).

We adopt a simple approach that trades linguistic annotation
fidelity for ease of automation, and consequently, scale.
For \datasetName, we consider an expanded vocabulary
beyond the 1,064 word vocabulary studied in~\cite{Albanie2020bsl1k}.
Concretely, to select a vocabulary set for a sign language recognition benchmark,
we first lemmatise each word in the subtitles
(this can be viewed as an approximation to mapping English
words to their corresponding glosses).
Next, we filter the candidate words to include only those
that appear in the training set amongst the mouthing annotations
with a confidence of 0.8 on at least 5 occasions. 
We then remove a small number of words for which we have found
(through human verification) that the mouthings consistently
failed to correspond to signs.
Specifically, we removed words for which:
(i) we had at least 15 spottings, and
(ii) annotators marked at least 95\% of the spottings as false
positives (i.e.\ the sign did not correspond to the lemmatised term).
Filtered words included terms like
\textit{therefore}, \textit{just} and \textit{if}.
Finally, we removed terms from the vocabulary that had no
verified instances and did not occur in the
vocabulary of BSLDict~\cite{momeni20watchread}.
We did not filter against 
    SignBank\footnote{\url{https://bslsignbank.ucl.ac.uk/}} as opposed to 
    the conservative vocabulary of BSL-1K\cite{Albanie2020bsl1k} since the lexicon of SignBank (2,016 words)
    is more restricted than BSLDict (9,283 words \& phrases).
The result of this filtering process was a set of \vocabTwoK words
(which includes a number of proper nouns)
that we take to constitute the sign language recognition vocabulary.
We expect this set of vocabulary to evolve and expand over time
    with better sign localisation methods, it may also be possible to potentially
    merge and split some categories.

\vspace{4pt}

\noindent \textbf{Automatic training and validation set of sign instances.} These automatic annotations are obtained through the methods described in Sec.~\ref{subsec:auto-annotation}.
Statistics for the different partitions of the training set from mouthing, dictionary and attention methods are shown in Fig.~\ref{fig:recognition-train}, as well as Tab.~\ref{tab:recognition-test-set}.
With a confidence score threshold of 0.5, the mouthing sign spottings yielded
707K and 15K annotations on the training and validation sets, respectively.
With a similarity score threshold of 0.7, the dictionary sign spottings yielded
5M training and 126K validation annotations. The attention sign spottings
(which do not require a threshold) yielded 434K training and 9K validation annotations.
We note that there exists a trade-off between the amount of annotations
and the level of noise. We therefore plot the distribution of each annotation
type according to their associated scores in Fig.~\ref{fig:recognition-train}.
We also show the portion of the data belonging to our vocabulary of \vocabTwoK signs.

\vspace{4pt}

\begin{figure}[t]
    \centering
    \includegraphics[width=.48\textwidth,trim={0cm 0cm 0cm 0cm},clip]{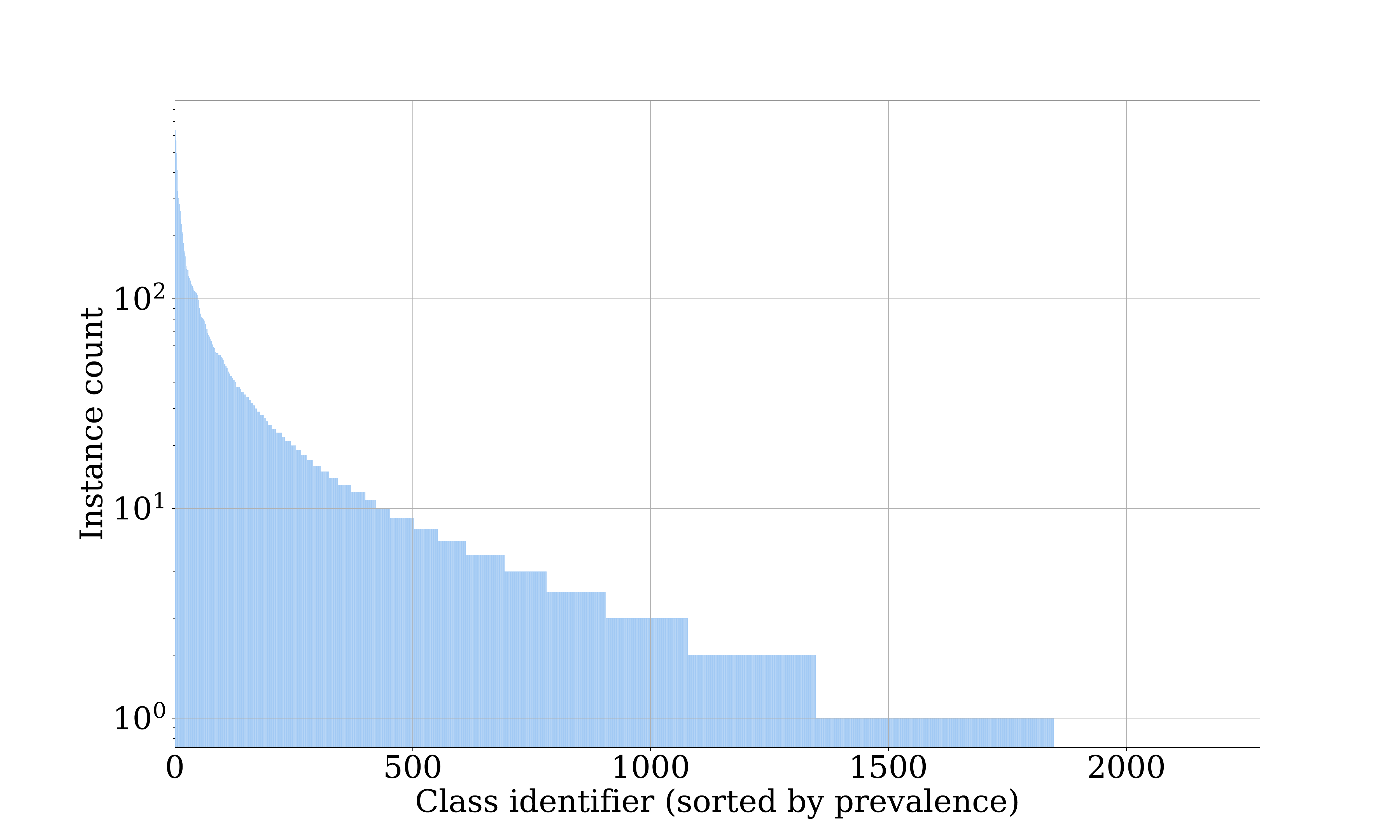}
    \caption{\textbf{Distribution of verified signs for the recognition test set.}
    As with real-world usage, the frequencies of annotated signs across
    the test set follow a power law distribution (note that the y axis uses a log scale).
    Here, class labels are sorted by prevalence along the x-axis for ease of visualisation. 
   }
    \label{fig:verified-sign-distribution}
\end{figure}

\noindent \textbf{Human-verified test set of sign instances.}
The human-verified sign annotations are obtained through the
process described in Sec.~\ref{subsec:manual-annotation}.
Statistics for the test set from verified
spottings
are shown in Tab.~\ref{tab:recognition-test-set}.
\testSign has a total of 25,045 verified sign instances
(9,263 from mouthings, 15,782 from dictionary spottings)
which span 1,849 elements of the \vocabTwoK vocabulary.
The distribution of annotations exhibits a power law,
visualised in Fig.~\ref{fig:verified-sign-distribution}. 
\begin{table}
    \setlength{\tabcolsep}{3pt}
    \centering
    \caption{
    \textbf{Splits for sign instances.}
    We report the total number of sign instances in the
    automatic training set (without thresholding the scores)
    and the human-verified test set.
    For each set, we show the amount of sign instance annotations
    with and without filtering to the vocabulary of \vocabTwoK,
    as well as the total effective vocabulary of words if we do not filter.
    }
    \resizebox{0.99\linewidth}{!}{
        \begin{tabular}{llrrr}
            \toprule
            Split & Spotting source & \#annots-2K & \#annots-full & vocab \\ %
            \midrule
            \trainM & mouthing & 502K & 707K & 22.3K \\
            \trainD & dictionary & 1.587M & 5.030M & 6.7K \\
            \trainA & attention & 286K & 434K & 1.4K \\
            \hdashline[0.5pt/5pt]
            \rule{0pt}{2ex} 
            \trainMDA & mouthing, dictionary, attention & 2.374M & 6.171M & 24.7K \\
            \midrule
            \midrule
            \valM & mouthing & 11K & 15K & 3.9K \\
            \valD & dictionary & 38K & 126K & 3.9K \\
            \valA & attention & 6K & 9K & 0.7K \\
            \hdashline[0.5pt/5pt]
            \rule{0pt}{2ex} 
            \valMDA & mouthing, dictionary, attention & 56K & 151K & 5.9K \\
            \midrule
            \midrule
            \testSign        & mouthing, dictionary & 25K & 25K & 1.8K \\
            \bottomrule
        \end{tabular}
    }
    \label{tab:recognition-test-set}
\end{table}

\subsection{\datasetName partitions for \textit{sentence} alignment
and translation evaluations}\label{subsec:splits-trans-align}

In order to develop methods for sign language sentence alignment
and translation,
we need aligned continuous signing segments and
corresponding English sentences.
We propose to make use of two levels of alignment:
(i) audio-aligned video-sentence alignments that have been
filtered using automatic spotting annotations
to select sentences that are likely to be reasonably well
aligned to the signing (these are available in large numbers);
(ii) manual video-sentence alignments 
(these are available in smaller numbers). 
\vspace{4pt}

\noindent \textbf{\SpottingFiltered signing video-sentence alignments.}
These correspond to video segments for which
an automatic sign instance annotation falls within the corresponding sentence timestamps (we restrict ourselves to annotations obtained from mouthings and dictionaries with confidence over 0.8 and use all annotations obtained through attention)
and the word matching the sign occurs in the sentence. This indicates a probable approximate alignment between the signing
video and corresponding sentence. For the sentence timestamps, we use the audio-aligned timestamps shifted by +2.7 seconds -- this is the average shift calculated between audio-aligned and signing-aligned sentences in our manual training set (\trainManual) described next. We define these splits as \trainSF, \valSF, \testSF.
These \textit{\spottingFiltered} alignments enable
large-scale training over multiple domains of discourse.

\begin{table}
    \setlength{\tabcolsep}{2pt}
    \centering
    \caption{
    \textbf{Aligned sentence-level subtitles.}
    Statistics summarising the \datasetName
    data for which manually aligned sentence-level
    subtitles (indicated with an $\textsc{H}$ subscript)
    and automatically ``spotting-filtered'' sentence-level subtitles (indicated with an $\textsc{SF}$ subscript) are available.
    See Sec.~\ref{subsec:manual-annotation} for a description of the annotation process and Sec.~\ref{subsec:splits-trans-align} for details on how these splits were constructed.
    $^\dagger$Note that \testManual consists of human-aligned sentences.
    $^*$Out-of-vocabulary (O-O-V) statistics reported w.r.t \trainManual and \trainSF.
    }
    \resizebox{1.0\linewidth}{!}{
        \begin{tabular}{lccccccc}
            \toprule
            Split        & Episodes   &  Signers & Sentences & Vocab. & O-O-V & Singletons & Duration \\
                         &            &          &           &        &       &            & (hours) \\
            \midrule
            
           \trainManual    & 16   & 16 & 9,168  & 8,906  & -     & 4,371 & 13 \\
           \valManual      & 4    & 3  & 1,973  & 3,528  & 1,127 & 1,837 & 3 \\
            \midrule
           \trainSF        & 1673 & 28 & 294,944 & 8,954 & -     & 23 & 1,236 \\
           \valSF          & 33   & 4  & 7,594   & 6,318 & 0 & 337 & 28 \\
            \midrule
           \testManualNoSpace$^{\dagger}$     & 36   & 3  & 20,870 & 13,641 & (\textsc{H}: 8,030, \textsc{SF}: 6,490)$^*$ & 5,604 & 31 \\
            \bottomrule
        \end{tabular}
    }
    \label{tab:aligned-splits}
\end{table}

\vspace{4pt}
\noindent \textbf{Manual signing video-sentence alignments.}
These manual sentence-level alignments are obtained through the
process described in Sec.~\ref{subsec:manual-annotation},
with statistics shown in Tab.~\ref{tab:aligned-splits}.
There is a total of 32K manually aligned sentences for a total
duration of 46 hours.
The training set episodes are chosen to maximise the number of
signers.
Given access only to the manual training set,
the number of out-of-vocabulary (OOV) words
is
1,127 words
for the validation set
and
8,030 words
for the test set.
The distribution of show topics for the different splits is shown in
Fig.~\ref{fig:manually-aligned-distribution}, with \textit{science \& nature} representing the largest proportion for all dataset splits (see Fig.~\ref{fig:pie_charts_topic}).
We define these splits as \trainManual, \valManual, \testManual.

\wip{
\begin{figure}[t]
    \centering
    \includegraphics[width=.48\textwidth,trim={0cm 0cm 0cm 0cm},clip]{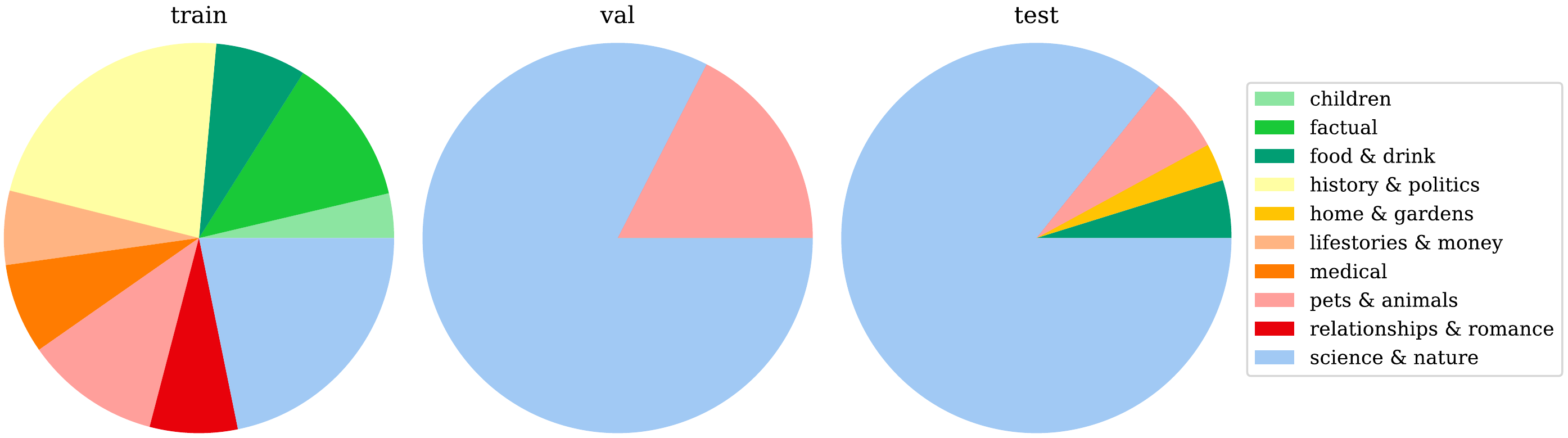}
    \caption{\textbf{Manual signing video-sentence alignment data divided into topics.}
    While \textit{science \& nature} represent the largest proportion of the validation and test set, the training set covers a broader range of themes.
     }
    \label{fig:manually-aligned-distribution}
\end{figure}
}

\section{Models and Implementation Details} \label{sec:models}

In this section, we describe the models employed for sign recognition
(Sec.~\ref{sec:models:recognition}),
sign language sentence alignment (Sec.~\ref{sec:models:alignment})
and sign language translation (Sec.~\ref{sec:models:translation}) baselines for the BOBSL dataset, as well as corresponding optimisation and implementation details.

\subsection{Sign recognition}
\label{sec:models:recognition}

\noindent \textbf{I3D architecture.}
We report sign recognition experiments using
a 3D convolutional model I3D~\cite{carreira2017quo}
which has been observed to perform well for
sign recognition~\cite{Joze19msasl,Li19wlasl}.
We report experiments for both RGB and
optical flow input streams~\cite{Simonyan2014TwoStreamCN}.
Flow is estimated with RAFT~\cite{teed2020raft}.
We initialise both models using Kinetics pretraining~\cite{carreira2017quo}
for each modality.
We input 16 consecutive
video frames at 25 fps,
corresponding to 0.64 seconds
(it has been noted in prior
work~\cite{buehler2009learning,pfister2013large,viitaniemi14}
that co-articulated (i.e. ``signs in context'') rather than 'isolated' 
signs typically exhibit a duration of
up to 13 frames, though this can vary significantly).
Video frames are processed at an initial spatial resolution
of $256 \times 256$ pixels:
this is then cropped to a $224 \times 224$
pixel square region and passed to the model
(during evaluation, this crop is taken from the centre of the frame;
during training, the frame is first resized isotropically along both
spatial dimensions by a scaling factor uniformly sampled between 0.875 and 1
and then centre-cropped).
Note, however, the flow is first estimated on the original
$444 \times 444$ resolution
(which we found produced qualitatively better results)
before being down-sampled to 256 pixels.
The input to the model is therefore of size $C \times 16 \times 224 \times 224$,
where $C=3$ for RGB and $C=2$ for flow.
During inference, the model produces class posterior
probabilities over a sliding window that spans the given
temporal interval of interest and the scores are
averaged to produce the final class predictions.

\vspace{4pt}

\noindent \textbf{\posetosign.}
We also experiment with a lightweight pose-based model,
following the approach described in~\cite{Albanie2020bsl1k}.
In particular, we extract pose estimates from each frame
using OpenPose~\cite{cao2019openpose}, 
yielding 70 facial, 15 upper-body and 21 per hand keypoints,
(we defer exploration of more powerful sequence-level 3D
pose-estimation techniques such as~\cite{jiang2021skeletor} to future work).
For the vast majority of frames, only one person is detected
(the BSL interpreter).
In the rare cases in which additional people are
visible (due to the content appearing behind the interpreter),
we select the pose for which keypoints have been estimated with
the greatest confidence to ensure that we process at most one
signer per frame.
The keypoints (consisting of $x,y$ coordinates and their
associated confidences) from 16 consecutive frames
are stacked to form a $3 \times 16 \times P$ pose image
(where $P$ denotes the total keypoint count per frame)
that is ingested by a 2D ResNet-18~\cite{he2016deep}
to perform sign recognition.

\vspace{4pt}

\noindent \textbf{Implementation details.}
We train on the \trainMD sign annotations that are above 0.8 confidence,
resulting in 426K training samples from the \vocabTwoK vocabulary.
For each sign annotation obtained from mouthing,
the model randomly samples a sequence of 16 contiguous frames
from a window covering 15 frames before the time associated with the annotation and 4 frames after,
i.e., $[-15, 4]$ around the mouthing peak. For dictionary annotations,
we use a window $[-3, 22]$ around the similarity peak. We set these intervals
based on preliminary experiments.

\vspace{4pt}

\noindent \textbf{Optimisation details.} For sign recognition experiments on \datasetName,
all models are trained for 25 epochs using SGD with momentum
(with a momentum value of 0.9), with a minibatch of size 4.
An initial learning rate of 0.01 is decayed by a factor of 10 after 20 epochs.
Scale and horizontal flip augmentations are applied
on the input video during training for all input modalities.
Colour augmentation is additionally during training on RGB input frames.

\subsection{Sign language sentence alignment}
\label{sec:models:alignment}

\noindent \textbf{SAT architecture.} We  use the Subtitle Aligner Transformer (SAT) model
from~\cite{Bull21} to temporally locate a text query corresponding to a sentence in a window of 
continuous signing. The encoder takes as input BERT token embeddings of the text query we wish to
align. The decoder takes as input a sequence of video features from a continuous sign language 
video segment extracted from an I3D model trained with \trainMD on sign classification
(described in Sec.~\ref{sec:models:recognition})
applied with a sliding window of stride 4. The decoder additionally takes as input a prior alignment: the shifted 
temporal boundaries of the audio-aligned text, i.e.\ +2.7 seconds where 2.7 is the average lag 
between the audio-aligned and annotated signing-aligned sentences in \trainManual. Using these 
inputs, the model outputs a vector of values between 0 and 1 of length equal to the length of 
video features. The temporal boundaries of sentences across the entire signing video under 
non-overlapping constraints is determined by maximising the sum of these output scores, using the
dynamic time warping algorithm \cite{myers1981comparative}.

\vspace{4pt}

\noindent \textbf{Implementation details.} We follow the procedure from~\cite{Bull21}, but pre-train the model using \trainSF in addition to \trainMD. We firstly pretrain SAT on word-level boundaries from \trainMD with confidence scores above 0.8, where we predict a 1-second interval centred at the automatic mouthing or dictionary sign instance annotation in a randomly chosen 20-second search window around the annotation. 
We do not input a prior alignment to the decoder. Secondly, we finetune this model using  the sentence-level boundaries from \trainSF, where we use random shifts of these sentence-level boundaries of up to 3 seconds as a prior alignment. Thirdly, we further finetune the model on sentence-level boundaries from \trainManual. We use 2.7-second shifted audio-aligned subtitles as a prior alignment, with additional random shifts of up to 2 seconds during training for data augmentation. When training on sentence-boundaries, we randomly select a search window of 20 seconds around the location of the prior alignment and filter to sentences longer than 0.5 seconds. We also randomly shuffle the words in 50\% of the sentences and drop 15\% of words during training as a data augmentation step. 

\vspace{4pt}

\noindent \textbf{Optimisation details.} We use the Adam optimiser with a batch size of 64. We train with a learning rate of $10^{-5}$ at the word-pretraining stage, $0.5\times10^{-5}$ at finetuning with sentence-level boundaries from \trainSF and $10^{-6}$ at finetuning with sentence-level boundaries from \trainManual. 
At the word pretraining stage, the model is trained over 22 epochs. During the full-sentence finetuning on \trainSF and \trainManual, the model is trained over 44 and 143 epochs respectively.

\subsection{Sign language translation}
\label{sec:models:translation}

\noindent \textbf{Transformer architecture.} We use a standard Transformer~\cite{Vaswani2017} encoder-decoder architecture that has been used in state-of-the-art work on sign language translation~\cite{camgoz2020sign} and signing video to token sequence prediction~\cite{Varol21}. Both the encoder and decoder consist of 2 attention layers, with 2 heads for each attention layer. The encoder input video consists of 1024-dimensional feature sequence extracted from an I3D model trained with \trainMD on sign classification for a vocabulary of \vocabTwoK signs
(described in Sec.~\ref{sec:models:recognition})
applied with a sliding window of stride 4.

\vspace{4pt}

\noindent \textbf{Implementation details.} We train the Transformer architecture with~\trainSF.
During training, the ground truth written English sequences are constructed by filtering
to a vocabulary of 9,163 words, obtained by selecting words which occur at least 50 times
in the training split subtitles. 
We note that we do not perform any lemmatising or stemming and we do not remove stop words
(as opposed to \cite{Varol21}). We subsequently filter to sentences with less than 30 words, giving us a total of 274K training samples (from 295K original samples in~\trainSF).

\vspace{4pt}

\noindent \textbf{Optimisation details.} We use the AdamW optimiser with a batch size of 64. We train for 70 epochs, with an initial learning rate of $10^{-10}$ reduced by a factor of 2 at 49th and 59th epochs.

\section{Experiments} \label{sec:experiments}

In this section, we provide baselines for the
tasks of sign language recognition (Sec.~\ref{sec:recognition-exps}),
sentence alignment (Sec.~\ref{sec:alignment-exps})
and translation (Sec.~\ref{sec:translation-exps}).

\subsection{Sign recognition}
\label{sec:recognition-exps}
\noindent \textbf{Evaluation protocol.} We evaluate on \testSign and report both top-1 and top-5 classification accuracy to better account for the extent of sign ambiguities that can be solved in context. We compute per-instance accuracy averaged over all test instances. We also measure per-class accuracy where we average over the sign categories present in the test set. The latter metric is helpful due to the unbalanced nature of the dataset.

\noindent \textbf{Baselines.}
We present results for three sign recognition baseline methods
on \testSign:
the simple \posetosign model,
together with \videotosign ingesting
either RGB or optical flow inputs.
We report the results in Tab.~\ref{table:verified-benchmark}.
We observe that of the three methods, RGB-\videotosign performs
best.
Nevertheless, there is considerable room for improvement
in performance, especially for the per-class accuracy,
underlining the challenging nature of the benchmark.

\begin{table}[h]
    \setlength{\tabcolsep}{9pt}
    \caption{
    \textbf{Sign recognition on \testSign.}
    A comparison of classification models that are trained on 426K automatic annotations
    in \trainMD
    for a vocabulary of \vocabTwoK signs. The evaluation is on the
    human-verified sign annotations.
    We observe that RGB-\videotosign performs best among individual modalities.
    }
    \centering
    \resizebox{0.99\linewidth}{!}{
        \begin{tabular}{lcccc}
            \toprule
            & \multicolumn{2}{c}{per-instance} & \multicolumn{2}{c}{per-class} \\
            Model & top-1 & top-5 & top-1 & top-5  \\
            \midrule
            2D \posetosign     & 61.8 & 82.1 & 30.6 & 56.6 \\
            Flow-\videotosign & 52.1 & 75.7 & 19.2 & 41.7 \\
            RGB-\videotosign  & 75.8 & 92.4 & 50.5 & 77.6 \\
            \bottomrule
        \end{tabular}
    }
    \label{table:verified-benchmark}
\end{table}

\subsection{Sign language sentence alignment}
\label{sec:alignment-exps}
\noindent \textbf{Evaluation protocol.} We evaluate on \testManual and measure frame accuracy and F1-score as in~\cite{Bull21}. For the F1-score, hits and misses of sentence alignment of sign language video are counted under three temporal overlap thresholds (0.1, 0.25, 0.5) between the predicted and ground truth signing-aligned sentences. For SAT, we select a search window of length 20 seconds centred around the shifted sentence location S$_{audio}^\text{+}$.

\noindent \textbf{Baselines.} We report the performance of baseline sign language alignment methods in Tab.~\ref{tab:alignment-baselines} on \testManual: (i)~the original audio-aligned subtitles (S$_{audio}$), (ii)~the shifted (by +2.7 seconds) audio-aligned subtitles (S$_{audio}^\text{+}$)
and (iii)~SAT model~\cite{Bull21}.
We observe that SAT performs best.
Results differ from those reported in~\cite{Bull21}, as we use sentences rather than subtitle texts. Moreover, we pretrain using word-level boundaries from \trainMD and finetune the model on sentence-level boundaries from \trainSF and \trainManual, rather than training only on BSL-1K and BSL-1K$_{aligned}$ and evaluating on the subtitle version of \testManual. 

\begin{table}[h]
    \caption{
    \textbf{Sign language alignment on \testManual.}
    We report baselines for sign language alignment on the 36 manually aligned episodes.
    We observe a significant improvement for SAT over the simpler baseline methods.
    }
    \centering
    \setlength{\tabcolsep}{6pt}
    \resizebox{0.9\linewidth}{!}{
    \begin{tabular}{lcccc}
        \toprule
        Method & frame-acc & F1@.10 & F1@.25 & F1@.50 \\
        \midrule
        S$_{audio}$ &  40.27 &  46.80 &  33.88 &  14.33 \\
        S$_{audio}^\text{+}$ & 62.33 & 73.01 & 64.28 & 44.75 \\
        \midrule
        SAT~\cite{Bull21} & \textbf{70.37} & \textbf{73.33} & \textbf{66.32} & \textbf{53.18} \\
        \bottomrule
    \end{tabular}
    }
    \label{tab:alignment-baselines}
\end{table}

\begin{table}%
    \setlength{\tabcolsep}{5pt}
    \caption{
    \textbf{Translation baseline on \testManual.
    } We
    report the results of training a Transformer translation model on \trainSF coarse video-sentence alignment.
    We observe that translation in such large-vocabulary, unconstrained settings remains very challenging.
    }
    \centering
    \resizebox{0.9\linewidth}{!}{
        \begin{tabular}{c|cccc}
            \toprule
          training \#samples  & Recall & BLEU-1 & BLEU-4  & ROUGE \\
            \midrule
             274K &  0.15  & 12.78 & 1.00 & 10.16\\
            \bottomrule
        \end{tabular}
    }
    \label{tab:translation-baselines}
\end{table}

\begin{figure}[h!]
    \begin{Verbatim}[fontsize=\scriptsize, frame=single, framerule=0.1mm, commandchars=\\\{\}]
    Example #1
    Ref: It's quite a journey especially if I get the bus.
    \textcolor{MidnightBlue}{Hyp: how long have you been in the bus now}
    
    Example #2
    Ref:  I'm also looking at migrating birds but from a 
    rather different angle.
    \textcolor{MidnightBlue}{Hyp: but the birds here make it look for the birds but}
    \textcolor{MidnightBlue}{this is the top}
    
    Example #3
    Ref:  It's hell of a difference yeah.
    \textcolor{MidnightBlue}{Hyp: it was like trying to be different to the world}
    
    Example #4
    Ref: But I think today I'm looking for something a bit 
    wilder.
    \textcolor{MidnightBlue}{Hyp: i really want to be a wild side really}
    
    Example #5
    Ref: I'm heading to a very special farm he set up here.
    \textcolor{MidnightBlue}{Hyp: so this is a farming farm that's a little bit special}
    
    \end{Verbatim}
    \caption{\textbf{Qualitative translation examples.}
    We show example target references, together with the translation hypotheses
    produced by the Transformer.
    While some target words are inferred correctly,
    the Transformer struggles to capture
    the meaning of the sentence.}
    \label{fig:qualitative-translation}
\end{figure}

\subsection{Sign language translation}
\label{sec:translation-exps}
\noindent \textbf{Evaluation protocol.} We evaluate on \testManual (without any vocabulary filtering of the ground truth sentences as in training) and measure recall of the model's predictions for each word by computing whether a word is correctly predicted,  averaging over the total number of words in all sequences. We also measure standard translation metrics such as BLEU-1, BLEU-4 and ROUGE.

\noindent \textbf{Baseline.}
We report the performance of our baseline sign language translation method
on \testManual in Tab.~\ref{tab:translation-baselines} and provide qualitative
examples in Fig.~\ref{fig:qualitative-translation}.
These results highlight the challenges of achieving large-vocabulary sign
language translation from videos to spoken language.
Given the significant room for improvement,
we hope this baseline underscores the need for
future sign language translation research in large-vocabulary scenarios.

\label{subsec:app:sparse}

\section{Opportunities and Limitations of the Data}
\label{sec:implications}

In this section we discuss some of the opportunities and limitations
of the data from several perspectives:
sign linguistics (Sec.~\ref{subsec:sign-linguistics}),
applications (Sec.~\ref{subsec:applications}),
annotator observations (Sec.~\ref{subsec:annotator-observations})
and dataset bias
(Sec.~\ref{subsec:ml-bias}).

\subsection{A sign linguistics perspective}
\label{subsec:sign-linguistics}

The availability of this dataset represents a positive advance
for enabling studies from a linguistics perspective. 
One challenge with existing technologically-focused
research on sign languages is that it has made use of
small databases, with few signers, limited content and limited naturalness.
The present dataset is large-scale, with a broad range of content,
and produced by signers of recognised high levels of proficiency. 
Nevertheless, there are limitations that should be recognised.
First among these is that although this is a relatively large dataset,
it includes only 39 signers, all using the same formal linguistic register,
and---because the signing is in the context of broadcast television---little
of the well-documented regional lexical variation in BSL~\cite{stamp2014lexical} is apparent.
Secondly,  all of the material is translated from English.
There is research evidence of systematic differences between interpreted
and non-interpreted language~\cite{dayter2019collocations}.
with evidence that differences in forms of language
are reduced in interpreted texts.
Finally, as an additional observation, we note that
there is some evidence of differences between the output of hearing
and deaf interpreters~\cite{stone2013interpreting},
which may manifest in the \datasetName data.

\subsection{Applications perspective}
\label{subsec:applications}

An important consideration when undertaking research in this area
is how useful/practical applications and outcomes can be produced
for deaf communities.
It should not be assumed that the views of hearing researchers
and deaf community members are fully aligned.
Consequently, to meet this objective, 
the involvement of deaf researchers and perspectives play a critical
role in defining target applications and outcomes.
Here we note two example applications that have been highlighted as
being of particular interest to deaf communities:
enabling indexing and efficient searchability of videos,
and providing sign-reading functionality comparable to voice-control
for interaction with various devices through applications like Siri and Alexa~\cite{bragg2019sign}.  
For the latter,
note that communication with virtual assistants through purely text-based interfaces have
significant practical limitations~\cite{glasser2020accessibility},
and even in cases when voices of DHH (Deaf and Hard of Hearing) individuals are identified as highly understandable
by professional speech pathologists and native hearing individuals, modern automatic speech
recognition systems struggle~\cite{glasser2019automatic}.
Prior work has shown that DHH
ASL signers preferred commands that were ASL-based over generic
gestures for virtual assistant interaction~\cite{rodolitz2019accessibility}.
By providing large-scale training data for computer vision models,
there is also an opportunity to improve automatic sign recognition to support
a signing interface to virtual assistants in BSL,
as well as to improve further applications such as \textit{search interfaces}
for sign language dictionaries,
for which retrieval quality correlates strongly with user
satisfaction~\cite{alonzo2019effect}.
Finally, we note that while the development of
improved sign language technology
has potential for positive impact,
it is valuable to be aware of historical context:
previous developments in sign language technology 
have struggled to deliver practical value~\cite{bragg2019sign,erard17}.
Sign language processing remains highly challenging,
and there remain significant research challenges to achieving
robust machine understanding of signing content~\cite{Yin2021IncludingSL}.

\subsection{Observations from the annotation process}
\label{subsec:annotator-observations}

During the process of constructing the dataset,
several observations arose from the annotation process
that provide useful additional context for working with \datasetName.
First, it was highlighted that it is frequently the case that not all words present in the
subtitles are captured by the signing of the BSL interpreter.
Instances when this occurs are tagged and provided as part of the manually aligned sentence 
annotations to support further analysis.
Second, it was noted that a small number of signs are used that would no longer
be considered appropriate in modern BSL.
These signs have been identified in the manually verified
spottings of the test set, and are excluded from evaluation.
However, we note that there are likely to be other occurrences of such signs
in the rest of the data.
We highlight this property to researchers working with the dataset,
with particular relevance for research that uses the data to train
sign language production models.

\subsection{Data bias}
\label{subsec:ml-bias}

While there are several promising research opportunities associated
with \datasetName, it is important to also recognise the limitations
of the dataset.
Here we note
factors that may have implications for the
generalisation of models trained on this data.
First, the data was gathered from TV broadcast footage:
consequently, the content of the signing reflects the content of TV shows,
rather than spontaneous, conversational signing.
A second consequence is that the distribution of interpreters
follows that of the original broadcasts, in which not all demographics are
equally represented.
A third consequence of using broadcast interpretations is that
the interpreters may choose not to convey information from the audio
stream that they consider to be redundant to the visual
stream of the footage.
Additional potential sources of bias stem from our use
of automatic annotation:
(1) First, the distribution of signs that were annotated
by spotting mouthings skew towards signs that are more
commonly associated with mouthing patterns,
as well as towards interpreters who sign with more
pronounced spoken components.
(2) Second, by constructing benchmark test sets for sign classification
through \textit{human verification of automatic sign proposals},
the distribution of test set signs will exhibit higher similarity
to the training set distribution than would be expected if the
test set was annotated without automatic proposals.
There is a trade-off here: 
our semi-automatic \textit{``propose and verify''} pipeline has the benefit of
significantly enhanced annotator efficiency
(enabling the creation of much larger and more comprehensive
test sets than would otherwise be possible).
However,
as a consequence of the bias introduced by the
\textit{propose and verify} pipeline,
researchers and practitioners should note the 
gap that remains between evaluation performance on
the \datasetName test sets
and expected performance on real world signing.
Noting these important caveats,
we nevertheless  hope that \datasetName forms a useful,
large-scale benchmark to spur progress
within the research community.

\section{Conclusion} \label{sec:conclusion}

We introduced \datasetName,
a large-scale dataset of British Sign Language.
We hope that this dataset will provide a useful
resource for researchers in the computer vision,
natural language processing and sign linguistics communities.

\ifCLASSOPTIONcompsoc
  \section*{Acknowledgments}
\else
  \subsection*{Acknowledgment}
\fi

The authors would like to thank Red Bee Media Ltd.\ and their BSL interpreters for supporting the research use of this data.
We would like to acknowledge the assistance of Andrew Brown
in preparing identity embeddings,
and Abhishek Dutta for his tireless support of the VIA annotation tool
development that was so essential for this project. We would also like to thank Ashish Thandavan, David Miguel Susano Pinto  and Ivan Johnson for their help in supporting the dataset release, and Necati Cihan Camg\"{o}z for helpful suggestions on dataset distribution.
This work was supported by EPSRC
grant ExTol, and a Royal Society Research Professorship.

\ifCLASSOPTIONcaptionsoff
  \newpage
\fi

\bibliographystyle{IEEEtran}
\bibliography{refs}

\begin{thebibliography}{10}
\providecommand{\url}[1]{#1}
\csname url@samestyle\endcsname
\providecommand{\newblock}{\relax}
\providecommand{\bibinfo}[2]{#2}
\providecommand{\BIBentrySTDinterwordspacing}{\spaceskip=0pt\relax}
\providecommand{\BIBentryALTinterwordstretchfactor}{4}
\providecommand{\BIBentryALTinterwordspacing}{\spaceskip=\fontdimen2\font plus
\BIBentryALTinterwordstretchfactor\fontdimen3\font minus
  \fontdimen4\font\relax}
\providecommand{\BIBforeignlanguage}[2]{{%
\expandafter\ifx\csname l@#1\endcsname\relax
\typeout{** WARNING: IEEEtran.bst: No hyphenation pattern has been}%
\typeout{** loaded for the language `#1'. Using the pattern for}%
\typeout{** the default language instead.}%
\else
\language=\csname l@#1\endcsname
\fi
#2}}
\providecommand{\BIBdecl}{\relax}
\BIBdecl

\bibitem{Albanie2020bsl1k}
S.~Albanie, G.~Varol, L.~Momeni, T.~Afouras, J.~S. Chung, N.~Fox, and
  A.~Zisserman, ``{BSL-1K}: {S}caling up co-articulated sign language
  recognition using mouthing cues,'' in \emph{European Conference on Computer
  Vision}, 2020.

\bibitem{sutton1999linguistics}
R.~Sutton-Spence and B.~Woll, \emph{The linguistics of {British Sign Language}:
  an introduction}.\hskip 1em plus 0.5em minus 0.4em\relax Cambridge University
  Press, 1999.

\bibitem{bragg2019sign}
D.~Bragg, O.~Koller, M.~Bellard, L.~Berke, P.~Boudreault, A.~Braffort,
  N.~Caselli, M.~Huenerfauth, H.~Kacorri, T.~Verhoef \emph{et~al.}, ``Sign
  language recognition, generation, and translation: An interdisciplinary
  perspective,'' in \emph{The 21st international ACM SIGACCESS conference on
  computers and accessibility}, 2019.

\bibitem{chai2014devisign}
X.~Chai, H.~Wang, and X.~Chen, ``The devisign large vocabulary of chinese sign
  language database and baseline evaluations,'' \emph{Technical report
  VIPL-TR-14-SLR-001. Key Lab of Intelligent Information Processing of Chinese
  Academy of Sciences (CAS), Institute of Computing Technology, CAS}, 2014.

\bibitem{csl500}
J.~Huang, W.~Zhou, H.~Li, and W.~Li, ``Attention-based {3D}-{CNN}s for
  large-vocabulary sign language recognition,'' \emph{IEEE Transactions on
  Circuits and Systems for Video Technology}, vol.~29, no.~9, pp. 2822--2832,
  2019.

\bibitem{asllvid2008}
V.~{Athitsos}, C.~{Neidle}, S.~{Sclaroff}, J.~{Nash}, A.~{Stefan}, {Quan Yuan},
  and A.~{Thangali}, ``The {American Sign Language} lexicon video dataset,'' in
  \emph{CVPRW}, 2008.

\bibitem{sehyr2021asl}
Z.~S. Sehyr, N.~Caselli, A.~M. Cohen-Goldberg, and K.~Emmorey, ``The {ASL-LEX}
  2.0 project: A database of lexical and phonological properties for 2,723
  signs in {American Sign Language},'' \emph{The Journal of Deaf Studies and
  Deaf Education}, vol.~26, no.~2, pp. 263--277, 2021.

\bibitem{Joze19msasl}
H.~R.~V. Joze and O.~Koller, ``{MS-ASL}: {A} large-scale data set and benchmark
  for understanding {American Sign Language},'' in \emph{British Machine Vision
  Conference (BMVC)}, 2019.

\bibitem{Li19wlasl}
D.~Li, C.~R. Opazo, X.~Yu, and H.~Li, ``Word-level deep sign language
  recognition from video: A new large-scale dataset and methods comparison,''
  in \emph{WACV}, 2019.

\bibitem{Momeni20b}
L.~Momeni, G.~Varol, S.~Albanie, T.~Afouras, and A.~Zisserman, ``Watch, read
  and lookup: Learning to spot signs from multiple supervisors,'' in
  \emph{Asian Conference on Computer Vision (ACCV)}, 2020.

\bibitem{ozdemir2020bosphorussign22k}
O.~{\"O}zdemir, A.~A. K{\i}nd{\i}ro{\u{g}}lu, N.~Cihan~Camgoz, and L.~Akarun,
  ``{BosphorusSign22k Sign Language Recognition Dataset},'' in
  \emph{Proceedings of the LREC2020 9th Workshop on the Representation and
  Processing of Sign Languages: Sign Language Resources in the Service of the
  Language Community, Technological Challenges and Application Perspectives},
  2020.

\bibitem{Sincan2020AUTSLAL}
O.~M. Sincan and H.~Keles, ``{AUTSL}: A large scale multi-modal {Turkish Sign
  Language} dataset and baseline methods,'' \emph{IEEE Access}, vol.~8, pp.
  181\,340--181\,355, 2020.

\bibitem{Sridhar2020INCLUDEAL}
A.~Sridhar, R.~G. Ganesan, P.~Kumar, and M.~M. Khapra, ``Include: A large scale
  dataset for indian sign language recognition,'' \emph{Proceedings of the 28th
  ACM International Conference on Multimedia}, 2020.

\bibitem{smile}
S.~Ebling, N.~Camgoz, P.~Braem, K.~Tissi, S.~Sidler-Miserez, S.~Stoll,
  S.~Hadfield, T.~Haug, R.~Bowden, S.~Tornay, M.~Razavi, and M.~Magimai-Doss,
  ``Smile swiss german sign language dataset,'' in \emph{LREC}, 2018.

\bibitem{viitaniemi14}
V.~Viitaniemi, T.~Jantunen, L.~Savolainen, M.~Karppa, and J.~Laaksonen, ``S-pot
  – a benchmark in spotting signs within continuous signing,'' in
  \emph{LREC}, 2014.

\bibitem{purdue06}
R.~B. Wilbur and A.~C. Kak, ``{Purdue RVL-SLLL} {A}merican sign language
  database,'' \emph{School of Electrical and Computer Engineering Technical
  Report, TR-06-12, Purdue University, W. Lafayette, IN 47906.}, 2006.

\bibitem{EfficientApproxJointTrackingRecognition}
P.~Dreuw, J.~Forster, T.~Deselaers, and H.~Ney, ``Efficient approximations to
  model-based joint tracking and recognition of continuous sign language,'' in
  \emph{IEEE International Conference on Automatic Face and Gesture
  Recognition}, 2008.

\bibitem{Duarte_CVPR2021}
A.~Duarte, S.~Palaskar, L.~Ventura, D.~Ghadiyaram, K.~DeHaan, F.~Metze,
  J.~Torres, and X.~Giro-i Nieto, ``{How2Sign: A Large-scale Multimodal Dataset
  for Continuous American Sign Language},'' in \emph{Conference on Computer
  Vision and Pattern Recognition (CVPR)}, 2021.

\bibitem{Huang2018VideobasedSL}
J.~Huang, W.~Zhou, Q.~Zhang, H.~Li, and W.~Li, ``Video-based sign language
  recognition without temporal segmentation,'' in \emph{AAAI}, 2018.

\bibitem{Zhou2021ImprovingSL}
H.~Zhou, W.~gang Zhou, W.~Qi, J.~Pu, and H.~Li, ``Improving sign language
  translation with monolingual data by sign back-translation,'' in \emph{IEEE
  Conference on Computer Vision and Pattern Recognition (CVPR)}, 2020.

\bibitem{signum2008}
U.~{von Agris}, M.~{Knorr}, and K.~{Kraiss}, ``The significance of facial
  features for automatic sign language recognition,'' in \emph{8th IEEE
  International Conference on Automatic Face Gesture Recognition}, 2008.

\bibitem{Koller15cslr}
O.~Koller, J.~Forster, and H.~Ney, ``Continuous sign language recognition:
  Towards large vocabulary statistical recognition systems handling multiple
  signers,'' \emph{Computer Vision and Image Understanding}, vol. 141, pp.
  108--125, 2015.

\bibitem{Camgoz18}
N.~C. Camgoz, S.~Hadfield, O.~Koller, H.~Ney, and R.~Bowden, ``Neural sign
  language translation,'' in \emph{Conference on Computer Vision and Pattern
  Recognition (CVPR)}, 2018.

\bibitem{ko2019neural}
S.-K. Ko, C.~J. Kim, H.~Jung, and C.~Cho, ``Neural sign language translation
  based on human keypoint estimation,'' 2019.

\bibitem{Kapoor2021TowardsAS}
P.~Kapoor, R.~Mukhopadhyay, S.~B. Hegde, V.~Namboodiri, and C.~V. Jawahar,
  ``Towards automatic speech to sign language generation,'' in
  \emph{INTERSPEECH}, 2021.

\bibitem{adaloglou2020comprehensive}
N.~Adaloglou, T.~Chatzis, I.~Papastratis, A.~Stergioulas, G.~T. Papadopoulos,
  V.~Zacharopoulou, G.~J. Xydopoulos, K.~Atzakas, D.~Papazachariou, and
  P.~Daras, ``A comprehensive study on sign language recognition methods,''
  \emph{arXiv preprint arXiv:2007.12530}, 2020.

\bibitem{camgoz2021content4all}
N.~C. Camgoz, B.~Saunders, G.~Rochette, M.~Giovanelli, G.~Inches,
  R.~Nachtrab-Ribback, and R.~Bowden, ``Content4all open research sign language
  translation datasets,'' \emph{arXiv preprint arXiv:2105.02351}, 2021.

\bibitem{schembri2013building}
A.~Schembri, J.~Fenlon, R.~Rentelis, S.~Reynolds, and K.~Cormier, ``Building
  the {B}ritish sign language corpus,'' \emph{Language Documentation \&
  Conservation}, vol.~7, pp. 136--154, 2013.

\bibitem{Varol21}
G.~Varol, L.~Momeni, S.~Albanie, T.~Afouras, and A.~Zisserman, ``Read and
  attend: Temporal localisation in sign language videos,'' in \emph{Conference
  on Computer Vision and Pattern Recognition (CVPR)}, 2021.

\bibitem{renz2021signtcn}
K.~Renz, N.~C. Stache, S.~Albanie, and G.~Varol, ``Sign language segmentation
  with temporal convolutional networks,'' in \emph{IEEE International
  Conference on Acoustics, Speech and Signal Processing (ICASSP)}, 2021.

\bibitem{renz2021sign}
K.~Renz, N.~C. Stache, N.~Fox, G.~Varol, and S.~Albanie, ``Sign segmentation
  with changepoint-modulated pseudo-labelling,'' in \emph{IEEE/CVF Conference
  on Computer Vision and Pattern Recognition Workshops (CVPRW)}, 2021.

\bibitem{Bull21}
H.~Bull, T.~Afouras, G.~Varol, S.~Albanie, L.~Momeni, and A.~Zisserman,
  ``Aligning subtitles in sign language videos,'' in \emph{International
  Conference on Computer Vision (ICCV)}, 2021.

\bibitem{Chung16a}
J.~S. Chung and A.~Zisserman, ``Out of time: automated lip sync in the wild,''
  in \emph{Workshop on Multi-view Lip-reading, ACCV}, 2016.

\bibitem{cao2019openpose}
Z.~Cao, G.~Hidalgo, T.~Simon, S.-E. Wei, and Y.~Sheikh, ``{OpenPose}: Realtime
  multi-person {2D} pose estimation using part affinity fields,'' \emph{IEEE
  Transactions on Pattern Analysis and Machine Intelligence}, vol.~43, no.~1,
  pp. 172--186, 2019.

\bibitem{deng2019retinaface}
J.~Deng, J.~Guo, Y.~Zhou, J.~Yu, I.~Kotsia, and S.~Zafeiriou, ``Retinaface:
  Single-stage dense face localisation in the wild,'' \emph{arXiv preprint
  arXiv:1905.00641}, 2019.

\bibitem{howard2017mobilenets}
A.~G. Howard, M.~Zhu, B.~Chen, D.~Kalenichenko, W.~Wang, T.~Weyand,
  M.~Andreetto, and H.~Adam, ``Mobilenets: Efficient convolutional neural
  networks for mobile vision applications,'' \emph{arXiv preprint
  arXiv:1704.04861}, 2017.

\bibitem{yang2016wider}
S.~Yang, P.~Luo, C.-C. Loy, and X.~Tang, ``Wider face: A face detection
  benchmark,'' in \emph{IEEE Conference on Computer Vision and Pattern
  Recognition (CVPR)}, 2016.

\bibitem{hu2018squeeze}
J.~Hu, L.~Shen, and G.~Sun, ``Squeeze-and-excitation networks,'' in \emph{IEEE
  Conference on Computer Vision and Pattern Recognition (CVPR)}, 2018.

\bibitem{scikit-learn}
F.~Pedregosa, G.~Varoquaux, A.~Gramfort, V.~Michel, B.~Thirion, O.~Grisel,
  M.~Blondel, P.~Prettenhofer, R.~Weiss, V.~Dubourg, J.~Vanderplas, A.~Passos,
  D.~Cournapeau, M.~Brucher, M.~Perrot, and E.~Duchesnay, ``Scikit-learn:
  Machine learning in {P}ython,'' \emph{Journal of Machine Learning Research},
  vol.~12, pp. 2825--2830, 2011.

\bibitem{woll2001u}
B.~Woll, ``The sign that dares to speak its name: echo* phonology in {B}ritish
  {S}ign {L}anguage ({BSL}),'' \emph{The Hands are the Head of the mouth: The
  Mouth as Articulator in Sign Languages, P. Boyes Braem \& R. Sutton-Spence
  (eds.), 87–98}, 2001.

\bibitem{sutton2007mouthings}
R.~Sutton-Spence, ``Mouthings and simultaneity in {British Sign Language},'' in
  \emph{Simultaneity in Signed Languages: From and Function}.\hskip 1em plus
  0.5em minus 0.4em\relax John Benjamins Publishing Company, 2007, pp.
  147--162.

\bibitem{flint2017text}
E.~Flint, E.~Ford, O.~Thomas, A.~Caines, and P.~Buttery, ``A text normalisation
  system for non-standard english words,'' in \emph{Proceedings of the 3rd
  Workshop on Noisy User-generated Text}, 2017, pp. 107--115.

\bibitem{cmu}
{Speech Group at Carnegie Mellon University}, ``{{CMU}} pronouncing
  dictionary,'' \url{http://www.speech.cs.cmu.edu/cgi-bin/cmudict}, 2014.

\bibitem{stafylakis2018zero}
T.~Stafylakis and G.~Tzimiropoulos, ``Zero-shot keyword spotting for visual
  speech recognition in-the-wild,'' in \emph{European Conference on Computer
  Vision (ECCV)}, 2018.

\bibitem{Momeni2020kws}
L.~Momeni, T.~Afouras, T.~Stafylakis, S.~Albanie, and A.~Zisserman, ``Seeing
  wake words: Audio-visual keyword spotting,'' \emph{arXiv}, 2020.

\bibitem{chung2016lip}
J.~S. Chung and A.~Zisserman, ``Lip reading in the wild,'' in \emph{Asian
  Conference on Computer Vision (ACCV)}.\hskip 1em plus 0.5em minus 0.4em\relax
  Springer, 2016.

\bibitem{chung2016signs}
------, ``Signs in time: Encoding human motion as a temporal image,''
  \emph{arXiv preprint arXiv:1608.02059}, 2016.

\bibitem{varol21bslattend}
G.~Varol, L.~Momeni, S.~Albanie, T.~Afouras, and A.~Zisserman, ``Read and
  attend: Temporal localisation in sign language videos,'' in \emph{Conference
  on Computer Vision and Pattern Recognition (CVPR)}, 2021.

\bibitem{Vaswani2017}
A.~Vaswani, N.~Shazeer, N.~Parmar, J.~Uszkoreit, L.~Jones, A.~N. Gomez,
  {\L}.~Kaiser, and I.~Polosukhin, ``Attention is all you need,'' in
  \emph{NeurIPS}, 2017.

\bibitem{jiang2021looking}
T.~Jiang, N.~C. Camgoz, and R.~Bowden, ``Looking for the signs: Identifying
  isolated sign instances in continuous video footage,'' \emph{IEEE
  International Conferene on Automatic Face and Gesture Recognition}, 2021.

\bibitem{dutta2019via}
A.~Dutta and A.~Zisserman, ``The via annotation software for images, audio and
  video,'' in \emph{Proceedings of the 27th ACM international conference on
  multimedia}, 2019, pp. 2276--2279.

\bibitem{johnston2010archive}
T.~Johnston, ``From archive to corpus: Transcription and annotation in the
  creation of signed language corpora,'' \emph{International journal of corpus
  linguistics}, vol.~15, no.~1, pp. 106--131, 2010.

\bibitem{momeni20watchread}
L.~Momeni, G.~Varol, S.~Albanie, T.~Afouras, and A.~Zisserman, ``Watch, read
  and lookup: learning to spot signs from multiple supervisors,'' in
  \emph{ACCV}, 2020.

\bibitem{carreira2017quo}
J.~Carreira and A.~Zisserman, ``Quo vadis, action recognition? a new model and
  the kinetics dataset,'' in \emph{IEEE Conference on Computer Vision and
  Pattern Recognition (CVPR)}, 2017.

\bibitem{Simonyan2014TwoStreamCN}
K.~Simonyan and A.~Zisserman, ``Two-stream convolutional networks for action
  recognition in videos,'' in \emph{NeurIPS}, 2014.

\bibitem{teed2020raft}
Z.~Teed and J.~Deng, ``Raft: Recurrent all-pairs field transforms for optical
  flow,'' in \emph{European Conference on Computer Vision (ECCV)}.\hskip 1em
  plus 0.5em minus 0.4em\relax Springer, 2020.

\bibitem{buehler2009learning}
P.~Buehler, A.~Zisserman, and M.~Everingham, ``Learning sign language by
  watching tv (using weakly aligned subtitles),'' in \emph{IEEE Conference on
  Computer Vision and Pattern Recognition (CVPR)}, 2009.

\bibitem{pfister2013large}
T.~Pfister, J.~Charles, and A.~Zisserman, ``Large-scale learning of sign
  language by watching tv (using co-occurrences),'' in \emph{British Machine
  Vision Conference (BMVC)}, 2013.

\bibitem{jiang2021skeletor}
T.~Jiang, N.~C. Camgoz, and R.~Bowden, ``Skeletor: Skeletal transformers for
  robust body-pose estimation,'' in \emph{IEEE/CVF Conference on Computer
  Vision and Pattern Recognition (CVPR)}, 2021.

\bibitem{he2016deep}
K.~He, X.~Zhang, S.~Ren, and J.~Sun, ``Deep residual learning for image
  recognition,'' in \emph{Conference on Computer Vision and Pattern Recognition
  (CVPR)}, 2016.

\bibitem{myers1981comparative}
C.~S. Myers and L.~R. Rabiner, ``A comparative study of several dynamic
  time-warping algorithms for connected-word recognition,'' \emph{Bell System
  Technical Journal}, vol.~60, no.~7, pp. 1389--1409, 1981.

\bibitem{camgoz2020sign}
N.~C. Camgoz, O.~Koller, S.~Hadfield, and R.~Bowden, ``Sign language
  transformers: Joint end-to-end sign language recognition and translation,''
  in \emph{IEEE Conference on Computer Vision and Pattern Recognition (CVPR)},
  2020.

\bibitem{stamp2014lexical}
R.~Stamp, A.~Schembri, J.~Fenlon, R.~Rentelis, B.~Woll, and K.~Cormier,
  ``Lexical variation and change in {British Sign Language},'' \emph{PLoS One},
  vol.~9, no.~4, p. e94053, 2014.

\bibitem{dayter2019collocations}
D.~Dayter, ``Collocations in non-interpreted and simultaneously interpreted
  english: a corpus study,'' in \emph{New empirical perspectives on translation
  and interpreting}.\hskip 1em plus 0.5em minus 0.4em\relax Routledge, 2019,
  pp. 67--91.

\bibitem{stone2013interpreting}
C.~Stone and D.~Russell, ``Interpreting in international sign: Decisions of
  deaf and non-deaf interpreters,'' 2013.

\bibitem{glasser2020accessibility}
A.~Glasser, V.~Mande, and M.~Huenerfauth, ``Accessibility for deaf and hard of
  hearing users: Sign language conversational user interfaces,'' in
  \emph{Proceedings of the 2nd Conference on Conversational User Interfaces},
  2020, pp. 1--3.

\bibitem{glasser2019automatic}
A.~Glasser, ``Automatic speech recognition services: deaf and hard-of-hearing
  usability,'' in \emph{Extended Abstracts of the 2019 CHI Conference on Human
  Factors in Computing Systems}, 2019, pp. 1--6.

\bibitem{rodolitz2019accessibility}
J.~Rodolitz, E.~Gambill, B.~Willis, C.~Vogler, and R.~Kushalnagar,
  ``Accessibility of voice-activated agents for people who are deaf or hard of
  hearing,'' \emph{Journal on Technology and Persons with Disabilities},
  vol.~7, pp. 144--156, 2019.

\bibitem{alonzo2019effect}
O.~Alonzo, A.~Glasser, and M.~Huenerfauth, ``Effect of automatic sign
  recognition performance on the usability of video-based search interfaces for
  sign language dictionaries,'' in \emph{The 21st International ACM SIGACCESS
  Conference on Computers and Accessibility}, 2019, pp. 56--67.

\bibitem{erard17}
M.~Erard, ``Why sign-language gloves don’t help deaf people,'' The Atlantic,
  \url{https://www.theatlantic.com/technology/archive/2017/11/why-sign-language-gloves-dont-help-deaf-people/545441/},
  2017.

\bibitem{Yin2021IncludingSL}
K.~Yin, A.~Moryossef, J.~A. Hochgesang, Y.~Goldberg, and M.~Alikhani,
  ``Including signed languages in natural language processing,'' \emph{ACL},
  2021.

\end{thebibliography}

\end{document}